\documentclass{article}

\usepackage{spconf}
\usepackage{amsmath}
\usepackage{graphicx}

\usepackage{booktabs}
\usepackage{multirow}
\usepackage{siunitx}
\usepackage[table]{xcolor}
\usepackage{tikz}
\usepackage{fvextra}
\usepackage{makecell}
\usepackage{fancyvrb}

\usepackage{xurl}
\usepackage{ragged2e}
\usepackage{enumitem}
\usepackage{eso-pic}

\usepackage{placeins}

\definecolor{mygreen}{HTML}{F6FFF0}
\definecolor{myred}{HTML}{FFF0F0}

\makeatletter
\let\savedmaketitle\maketitle
\let\saved@maketitle\@maketitle
\makeatother

\newcommand{\SetPaperTitle}{%
    \title{Toward Semantic-Agnostic and Shape-Aware\\Vision-Language Segmentation Models}
    \name{Corentin Seutin$^{\;1,\;2}$ \;  Mohamed Amine Ettaki$^{\;1}$ \; Michaël Clément$^{\;1}$ \; Pierrick Coupé$^{\;1}$ \; Rémi Giraud$^{\;2}$}
    \address{$^{\;1}$Univ. Bordeaux, CNRS, Bordeaux INP, LaBRI, UMR 5800, France
    \and 
    $^{\;2}$Univ. Bordeaux, CNRS, Bordeaux INP, IMS, UMR 5218, France
    }
}

\newcommand{\SetPaperTitleSupMat}{%
    \title{Toward Semantic-Agnostic and Shape-Aware\\Vision-Language Segmentation Models\\
    --- Supplementary Material ---}
    \name{Corentin Seutin$^{\;1,\;2}$ \;  Mohamed Amine Ettaki$^{\;1}$ \; Michaël Clément$^{\;1}$ \; Pierrick Coupé$^{\;1}$ \; Rémi Giraud$^{\;2}$}
    \address{$^{\;1}$Univ. Bordeaux, CNRS, Bordeaux INP, LaBRI, UMR 5800, France
    \and 
    $^{\;2}$Univ. Bordeaux, CNRS, Bordeaux INP, IMS, UMR 5218, France
    }
}

\begin{document}
\topmargin=0mm

\SetPaperTitle
\maketitle

\AddToShipoutPictureFG*{
  \AtPageLowerLeft{
    \put(48,30){
      \parbox[b]{7.0in}{
        \scriptsize
        \textit{
        Copyright 2026 IEEE. Published in 2026 IEEE International Conference on Image Processing (ICIP), scheduled for 13-17 September 2026 in Tampere, Finland. Personal use of this material is permitted. Permission from IEEE must be obtained for all other uses, in any current or future media, including reprinting/republishing this material for advertising or promotional purposes, creating new collective works, for resale or redistribution to servers or lists, or reuse of any copyrighted component of this work in other works.
        }
    }
    }
  }
}

\bigskip

\begin{abstract}
Vision–language segmentation models have recently achieved strong performance by leveraging high-level semantic object categories expressed in natural language. However, this semantic dependence limits their ability to reason about intrinsic visual properties such as shape, geometry, or texture, which are essential in many real-world applications.
In this work, we introduce Semantic-Agnostic aNd Shape-Aware (SANSA) segmentation, a new paradigm that requires segmentation models to operate solely from non-semantic textual descriptions.
To this end, we propose two strategies to generate SANSA segmentation prompts 
based on either dictionary constraints or example guidance, both generating semantic-agnostic textual descriptions.
These prompts are then used to finetune segmentation models under semantic-agnostic supervision.
Experiments show that finetuning on SANSA prompts 
yields up to a 20\% mIoU improvement on this new segmentation task, compared to pretrained state-of-the-art models, 
while maintaining strong performance on standard semantic prompts. 
These results highlight the importance of low- and mid-level visual reasoning 
for improving the generalization and controllability of vision–language segmentation models.
Code and datasets are available here: \small{\url{github.com/CorentinSeutin/SANSA-ICIP-2026}}. 
\end{abstract}


\section{Introduction}
\label{sec:intro}

Image segmentation has seen major advances with the advent of deep learning, first driven by convolutional neural networks and later by Transformer-based \cite{vaswani2017attention} 
architectures such as Vision Transformers (ViTs) \cite{dosovitskiy2021image}, which demonstrated a high efficiency in extracting rich visual representations.
More recently, the emergence of foundation models such as Segment Anything Model (SAM) \cite{kirillov2023segment} marked a turning point: trained on massive datasets, these models offer robust generalization and enable segmentation guided by interactive prompts, paving the way for more flexible uses. 

In parallel, the emergence 
of large language models (LLMs) has transformed the way we interact with 
such systems,
thanks to their ability to reason and interpret complex instructions in natural language. 
Their extension to the visual domain has led to 
Vision-Language Models (VLMs)
\cite{carion2025sam,yang2022lavt,xia2024gsva,rasheed2024glamm,chen2024sam4mllm}, which combine image encoders,  
producing LLM-compatible tokens,
pretrained LLMs, 
and token decoders to jointly analyze text and images.

\begin{figure}[t]
 \centering
 {\footnotesize
    \begin{tabular}{@{\hspace{0mm}}cc@{\hspace{0mm}}}     
\includegraphics[width=4.0cm]{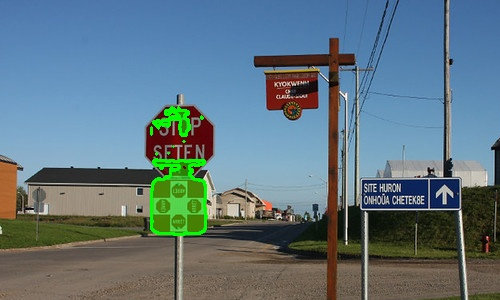}  &
      \includegraphics[width=4.0cm]{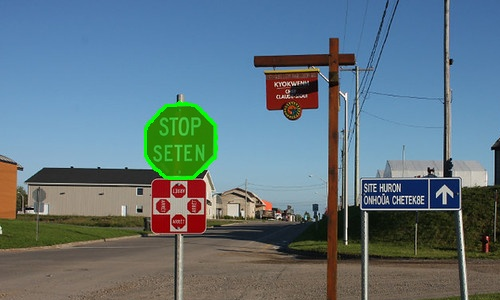}\\
     {(a) LISA \cite{lai2024lisa}} & (b) SANSA (ours)\\[-1.25ex]
     \end{tabular}
     }
    \caption{
        \textbf{Semantic-Agnostic aNd Shape-Aware (SANSA) segmentation example.}
        Comparison between (a) LISA~\cite{lai2024lisa}, and (b) our proposed finetuned SANSA model (EXSP). 
        Prompt: ``\texttt{\small Segment the red octagonal object}''.  \\[-3.75ex]
    }
    \label{fig:teaser}
\end{figure}

\begin{figure*}[t]
  \centering
\includegraphics[width=\textwidth]{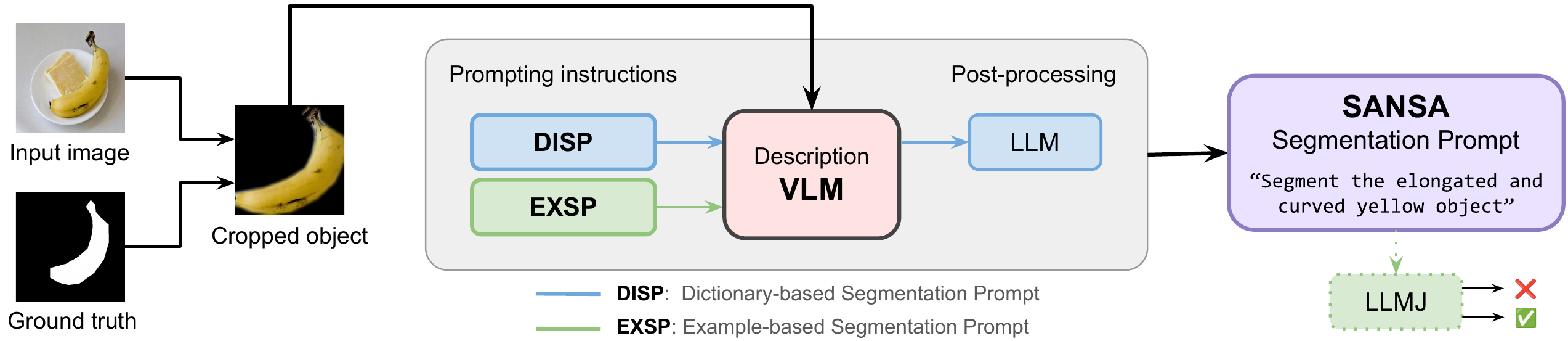}\\[-1.25ex]
\caption{
    \textbf{Semantic-agnostic and shape-aware (SANSA) segmentation prompt generation pipeline.} An object is cropped using its mask and fed to a description VLM.
    A semantic-agnostic object description is generated with either DISP or EXSP strategy. DISP descriptions 
    {are strictly constrained to a semantic-agnostic predefined vocabulary} and are just reformulated with LLM post-processing, while EXSP descriptions {may contain semantic words that} can be filtered by a LLM-as-a-judge (LLMJ).}
\label{fig:intro}
\end{figure*}

This paradigm has recently been applied to segmentation in approaches such as GSVA~\cite{xia2024gsva}, LISA~\cite{lai2024lisa}, or SegLLM~\cite{wang2025segllm}, where the task essentially consists of following a textual instruction usually based on an object category. 
These new segmentation assistants rely on complex multimodal architectures, but offer great flexibility to respond to a variety of queries.
Nevertheless, these models are still trained and validated on semantically annotated datasets such as COCO~\cite{lin2014microsoft} or LVIS~\cite{gupta2019lvis}, where masks correspond to high-level object classes. 
Therefore, they primarily learn to align text with semantic categories, rather than to reason about lower-level intrinsic properties of the images, such as shape, color, texture, or geometric structure.
This semantic framework limits the ability of such models, which can struggle to handle even simple non-semantic queries.
As shown in Fig.~\ref{fig:teaser}(a), the state-of-the-art LISA model~\cite{lai2024lisa} fails to segment the stop traffic sign when it is described by shape and color. 

Hence, VLMs may show limited generalization in non-semantic segmentation tasks.
While shape information has been shown to improve generalization in CNN models~\cite{zhang2024convolutional}, the role of non-semantic cues in VLMs remains largely unexplored.
Moreover, many applications require precise control over segmentations to satisfy properties such as the number of connected components or anatomical constraints~\cite{rouge2024ccdice}.

\subsubsection*{Contributions}

In this work, we introduce a new segmentation paradigm, Semantic-Agnostic aNd Shape-Aware (SANSA) segmentation, which aims to segment objects solely based on their appearance, including shape, color, texture, and spatial properties, as described in text queries. 
This paradigm enables a flexible and generalizable form of segmentation that does not rely on predefined object categories, subcategories, or any other high-level semantic information.

Consequently, existing vision–language models (VLMs) need to be adapted to tackle the SANSA task.
However, to the best of our knowledge, there exists no prompt set specifically designed to provide semantic-agnostic descriptions.
While manual prompt design is possible,
it does not scale to large datasets, making automatic generation with VLMs a natural alternative.
We therefore propose two strategies using a description VLM for constructing prompts suitable for SANSA segmentation: dictionary-based segmentation prompts (DISP) and 
example-based segmentation prompts (EXSP).
We then finetune a state-of-the-art segmentation VLM (LISA \cite{lai2024lisa}) on our SANSA prompts and demonstrate improved segmentation accuracy performance over pretrained models (see Fig~\ref{fig:teaser}(b)).

In summary, our contributions are as follows: \vspace{-0.2cm}
\begin{itemize}
  \setlength\itemsep{0em}
    \item We introduce SANSA segmentation, requiring models to segment objects based solely on their visual properties, without using any semantic information.
    \item We propose two strategies 
    to generate SANSA segmentation prompts containing only semantic-agnostic visual properties.
    \item We leverage these prompts to finetune a state-of-the-art VLM and demonstrate substantial improvements over pretrained models, highlighting enhanced generalization in semantic-agnostic segmentation.
\end{itemize}

\section{Shape-Aware and Semantic-Agnostic segmentation}
\label{sec:sansa-seg}

In this section, we describe our proposed Semantic-Agnostic aNd Shape-Aware (SANSA) segmentation paradigm with VLMs.
Our methodology consists of two main stages.
First, we introduce a pipeline for generating semantic-agnostic object descriptions using a first VLM in a constrained image captioning setting. 
These descriptions, referred to as SANSA segmentation prompts,
should contain only perceptual features such as shape, color, texture, and spatial properties, without including any semantic 
information.
Second, these 
prompts are used to finetune a second VLM in a text-based object segmentation setting, enabling it to segment objects guided solely by perceptual features.

\subsection{SANSA Segmentation Prompts}
\label{sec:sansa-seg:prompts}

The goal of this first stage is to automate the generation of semantic-agnostic object segmentation prompts.
To this end, we rely on a first description VLM (InternVL 2.5~\cite{chen2024expanding}), where the model is provided with an image crop of the target object and instructed to describe it.
This task is inherently challenging, as image captioning VLMs exhibit strong biases toward producing semantic or category-level information, even when explicitly instructed to avoid it (see supp. mat.).
To mitigate this issue, we introduce two complementary strategies designed to either constrain or guide the generation process toward semantic-agnostic descriptions.
An overview of this SANSA segmentation prompts generation pipeline is shown in Fig.~\ref{fig:intro}.
Finally, note that we consider as \textit{semantic}
all words that contain any high-level information about an object class or its subparts, that can be used for semantic learning.

\medskip\noindent\textbf{Object Cropping.}
In our pipeline, each target object is cropped directly using its binary mask, producing an image that contains only the object.
This masked object is then provided as input to the description VLM along with a prompt instructing it to generate a semantic-agnostic description of the object (see strategies presented hereafter).
This step is necessary because when the full image (or the object's bounding box) is provided, the description VLM tends to incorporate semantic information from the surrounding context into its description, even when explicitly instructed to avoid semantic cues (see supp. mat.). Cropping the object with its mask restricts the visual context, reducing the likelihood of unintended semantic references in the generated descriptions.

\medskip\noindent\textbf{DISP: Dictionary-based Segmentation Prompts.}
The idea of this first strategy is to constrain the 
VLM's output by forcing it to generate tokens exclusively from a predefined dictionary, thereby completely preventing the use of semantic words.
The allowed words 
include various semantic-agnostic attributes such as colors, textures, shapes, patterns, lighting, and connector words.
Hence, we use a relatively simple instruction to ask the VLM to generate a textual description
(see dictionary and prompt instructions in supp. mat.).

\begin{figure}
\centering\includegraphics[width=0.99\columnwidth]{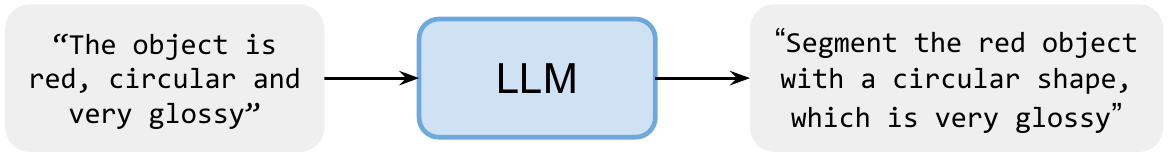}\\[-1ex]
\caption{LLM post-processing reformulates DISP object descriptions using an external LLM, producing more fluent and natural segmentation prompts while strictly preserving the original semantic-agnostic and perceptual information.}
\label{fig:LLM}
\end{figure}

We further refine the DISP descriptions using an external LLM (Mistral 7B~\cite{jiang2023mistral7b}).
This LLM is asked to reformulate each dictionary-constrained description into a more natural segmentation prompt, without adding any new information, while strictly preserving the semantic-agnostic and perceptual constraints of the original description.
This post-processing step produces prompts that are more fluent and readable, with higher resemblance to what a human annotator might write.
Fig.~\ref{fig:LLM} illustrates this LLM post-processing.

\medskip\noindent\textbf{EXSP: Example-based Segmentation Prompts.}
\label{sec:sansa-seg:EXSP}
This second strategy relies on providing the description VLM with a set of positive and negative examples to guide it in generating semantic-agnostic object descriptions.
Contrary to the first strategy limited by a dictionary, the idea behind this approach is to teach the model what types of descriptions are acceptable (positive examples without semantic) and what should be avoided (negative examples containing semantic words), without explicitly constraining the vocabulary.
This strategy may allow for greater variability and naturalness in the generated descriptions, producing outputs that are closer to human-written prompts. 
Yet, the instruction prompt, given examples and used VLM must be chosen cautiously. 
Otherwise the generated description is very likely to contain semantic information due to the biases of VLMs toward
producing semantic information (see supp. mat.).

\begin{figure}
\centering
\includegraphics[width=0.95\columnwidth]{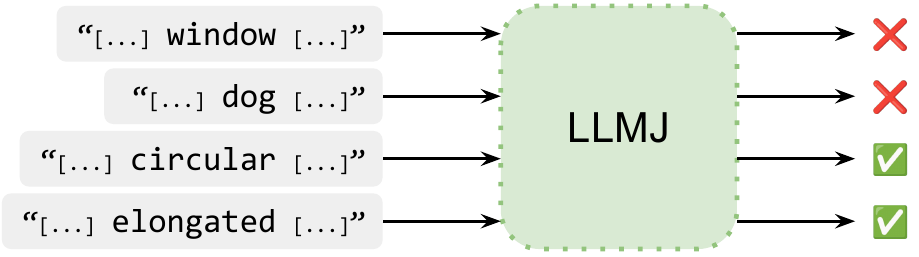}\\[-1.5ex]
\caption{
LLMJ analyzes EXSP object descriptions and filters out the ones containing semantic.
Note that the LLM is not provided a list of allowed words.
}
\label{fig:LLMJ}
\end{figure}

\medskip\noindent\textbf{LLMJ filtering for EXSP.}
To prevent the unintended inclusion of semantic information in the unconstrained EXSP strategy, we optionally apply a LLM-as-a-judge (LLMJ) filtering step.
In this approach, the description VLM (InternVL 2.5~\cite{chen2024expanding}) is tasked with classifying each EXSP description as either semantic or semantic-agnostic.
Note that only text is used, without the corresponding image.
Descriptions identified as containing semantic are then removed from the dataset (see Fig.~\ref{fig:LLMJ}).
The prompt used for LLMJ is given in supp. mat. and the robustness of this filtering is analyzed in Sec.~\ref{sec:experiments}. 

\subsection{Training for SANSA segmentation}
\label{sec:sansa-seg:training}

\begin{figure}
\centering
\includegraphics[width=0.99\columnwidth]{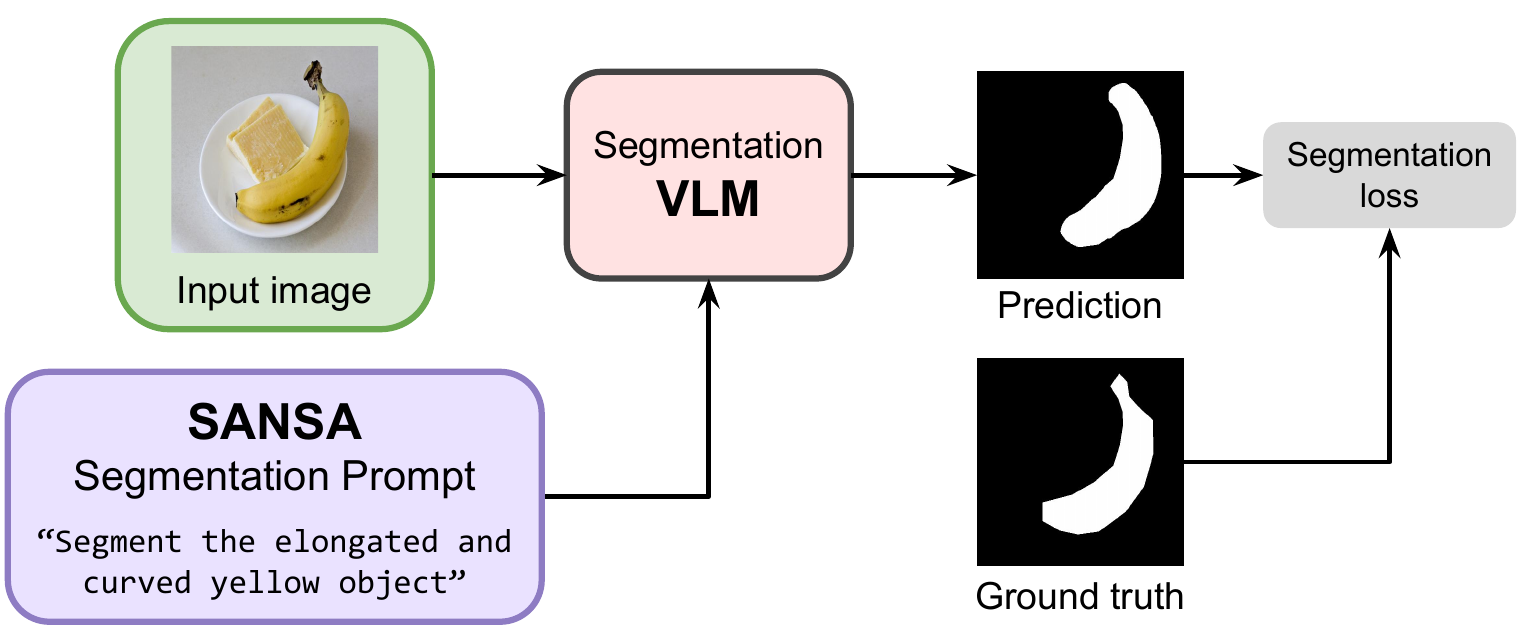}
\caption{\textbf{SANSA segmentation training pipeline.} SANSA prompts and input images are given to a segmentation VLM. The segmentation loss is computed on the prediction and the ground truth.}
\label{fig:training}
\end{figure}

As illustrated in Fig.~\ref{fig:training}, we use our processed SANSA prompts and their associated images to finetune a segmentation VLM.

\medskip\noindent\textbf{VLM Finetuning.}
We use LISA 7B \cite{lai2024lisa} as the pretrained segmentation VLM, as it can process long textual prompts to segment images.
Indeed, the SANSA paradigm requires significantly longer segmentation prompts, since explicit semantic labels such as classes or categories are not allowed.
This makes many other segmentation models, including SAM~\cite{kirillov2023segment} or LAVT \cite{yang2022lavt}, unsuitable for our task. 

Following LISA training, we employ LoRA~\cite{hu2022lora}, a lightweight adapter for finetuning large models.
Specifically, we use LoRA to finetune LLaVA~\cite{liu2023visual}, the multimodal model responsible for generating the segmentation token in LISA given an image-prompt pair.
We also train the vision decoder, which takes this token and produces the final segmentation. 
We finetune the segmentation model on the proposed DISP and EXSP segmentation prompts sets, 
enabling the model to segment the target object in an image based on its corresponding SANSA segmentation prompt.

While the EXSP strategy may occasionally include semantic words, this has little impact on finetuning as most prompts remain semantic agnostic, and the presence of some semantic cues may help the VLM retain its semantic segmentation capabilities.
Accordingly, we apply LLMJ filtering only at test time
to more accurately evaluate semantic-agnostic segmentation performance.

\medskip\noindent\textbf
{Segmentation loss.}
We use the same segmentation loss as LISA~\cite{lai2024lisa}, which is a weighted combination of pixelwise binary cross entropy (BCE) loss and Dice loss: 
\begin{equation}
\mathcal{L} = \lambda_1 \, \text{BCE}(\hat{Y}, Y) + \lambda_2 \, \text{Dice}(\hat{Y}, Y), 
\end{equation}
with $\lambda_1=0.25$ and $\lambda_2=1$, and $\hat{Y}$ the set of pixels predicted as object and $Y$ the corresponding ground truth.

\section{Experiments}
\label{sec:experiments}

\subsection{Experimental Setting}
\label{sec:experiments:settings}

\medskip\noindent\textbf{Datasets.}
To finetune models for the SANSA segmentation task, we use a subset of COCO comprising 10k images across the 80 categories (125 images per category).
This dataset is split into 8k images for training and 2k images for validation.
For test, we use a subset of the COCO val dataset, comprising 2k images (25 images per category).
We compute the SANSA segmentation prompts (both DISP and EXSP strategies) for all images of this dataset.
In addition, we introduce a Human Prompt (HP) dataset, consisting of 160 images sampled from our test set (2 images per category).
For these images, we manually produced realistic semantic-agnostic segmentation prompts to evaluate the generalization capacity of SANSA models in real-world scenarios with human prompts.

\medskip\noindent\textbf{Training Details.} We train on one NVIDIA RTX 6000 Ada GPU with 48 GB of VRAM, using a batch size of 2 and a learning rate of 0.0001.
Finetuning usually completes in 15 to 25 epochs with early stopping set to a patience of 5, requiring roughly 15 to 25 hours.

\medskip\noindent\textbf{Evaluation Metric.}
We evaluate the quality of segmentation with standard mean Intersection-over-Union (mIoU), computed as the average per-sample IoU:
{
\begin{equation}
    \text{mIoU} = \frac{1}{N} \sum_{n=1}^{N} \frac{|\hat{Y}_n \cap Y_n|}{|\hat{Y}_n \cup Y_n|} ,
\end{equation}
}

\noindent where $N$ is the total number of images, $\hat{Y}_n$ is {the set of pixels predicted for object $n$} and $Y_n$ is the corresponding ground truth.

\medskip\noindent\textbf{Comparative Methods.}
We evaluate our SANSA segmentation models against several recent segmentation VLMs.
LISA 7B~\cite{lai2024lisa} serves as our main baseline, as it is the model we finetune from.
We also consider GSVA 7B~\cite{xia2024gsva} and PolyFormer~\cite{liu2023polyformer}.
Note that other image-text segmentation models such as SAM~\cite{kirillov2023segment} or LAVT~\cite{yang2022lavt} are not included because their text input is restricted to a limited number of tokens, preventing them from efficiently processing full SANSA prompts.

\subsection{Experimental Results}
\label{sec:experiments:results}

\begin{table}[t]
    \caption{\textbf{Quantitative comparison (mIoU (\%))} of state-of-the-art pretrained segmentation VLMs and our SANSA finetuned models (8k training - 2k test data). 
   We report evaluation on all corresponding test sets and EXSP test set filtered by LLMJ, \emph{i.e.}, 
   more likely to respect SANSA constraints.
   The models are also evaluated on 
   high quality Human Prompts (HP).
    }
    \vspace{0.2cm}
    \renewcommand{\arraystretch}{1.25}
    \newcommand{\sepp}{2mm}
    \centering
    {\scriptsize
        \begin{tabular}{@{\hspace{0mm}}l@{\hspace{2mm}}l@{\hspace{2mm}}c@{\hspace{\sepp}}c@{\hspace{\sepp}}c@{\hspace{\sepp}}c@{\hspace{\sepp}}|c|@{\hspace{1mm}}|@{\hspace{1mm}}c@{\hspace{1mm}}}
        & & \multicolumn{6}{@{\hspace{0mm}}c@{\hspace{0mm}}}{Test sets (number of images)} \\
        \cline{3-8}
         & & \hspace{-0.75cm}\text{DISP w/o LLM} & \text{DISP} & \text{EXSP} & \text{EXSP+LLMJ} & Average 
         & \text{HP}  \\[-0.75ex]
         & &  (2k) & (2k) & (2k) & (1.5k) & (7.5k) &  (160) \\
        \toprule
        \multirow{3}{*}{\rotatebox{90}{Pretrained}}  
        & \text{GSVA 
        \cite{xia2024gsva}} & \text{15.61} & \text{15.28} & \text{24.68} & \text{15.61} & \text{17.80} & \text{25.35}  \\
        & \text{PolyFormer \cite{liu2023polyformer}} & \text{21.32} & \text{22.39} & \text{26.00} & \text{25.49} & \text{23.80} & \text{25.20} \\
        & \text{LISA 
        \cite{lai2024lisa}} & \text{17.98} & \text{19.80} & \text{27.72} & \text{26.41}& \text{22.98} & \text{25.48}  \\
        \midrule
         \multirow{3}{*}{\rotatebox{90}{Finetuned}} & DISP w/o LLM & \textbf{38.08} & \text{32.13} & \text{34.39} & \text{35.78} &  \text{35.10} & \text{37.68}  \\
        & {{DISP}} & \text{36.79} & \textbf{36.76} & \underline{36.42} & \underline{37.47}  & \underline{36.86}& \underline{39.47} \\
        & \text{EXSP} & \underline{37.77} & \underline{34.06} & \textbf{39.39} & \textbf{38.63} &  \textbf{37.46}&\textbf{42.98} \\
        \bottomrule
        \end{tabular}
        }
    \label{table:sansa-results}
\end{table}

\newcommand{\www}{0.23\textwidth}
\newcommand{\hhh}{0.17\textwidth}
\begin{figure*}[t]
      \centering
      {\footnotesize
      \begin{tabular}{@{\hspace{0mm}}c@{\hspace{4mm}}c@{\hspace{2mm}}c@{\hspace{2mm}}c@{\hspace{2mm}}c@{\hspace{0mm}}}
      \includegraphics[width=\www,height=\hhh]{imgs/paper-figs/qualitative-imgs4/000000066771-cropped.png}&
      \includegraphics[width=\www,height=\hhh]{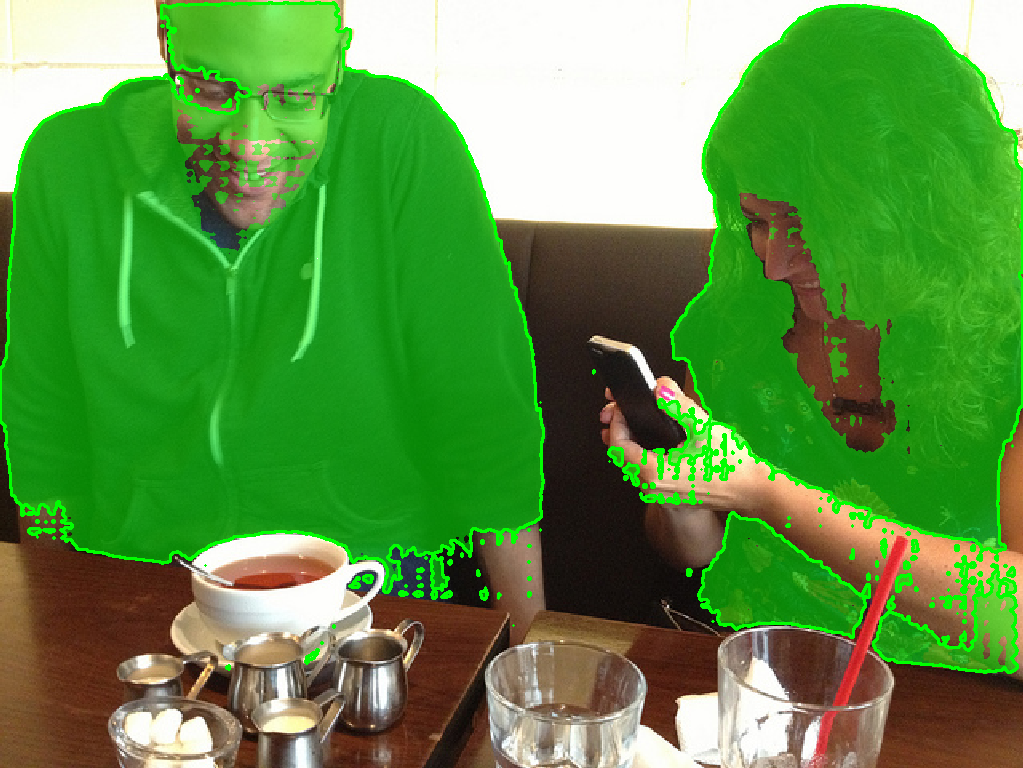}&
      \includegraphics[width=\www,height=\hhh]{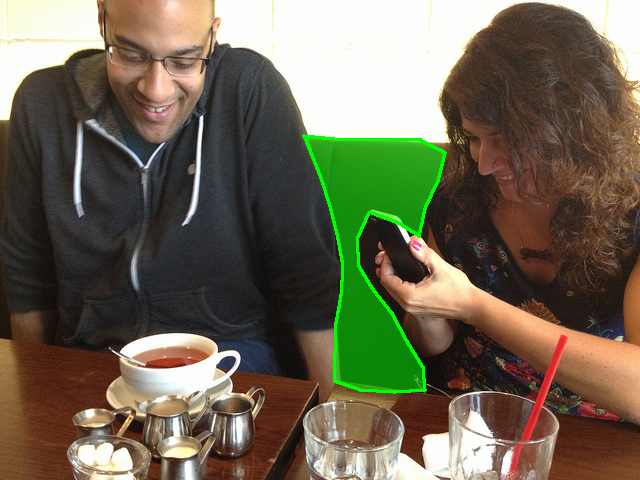}&
      \includegraphics[width=\www,height=\hhh]{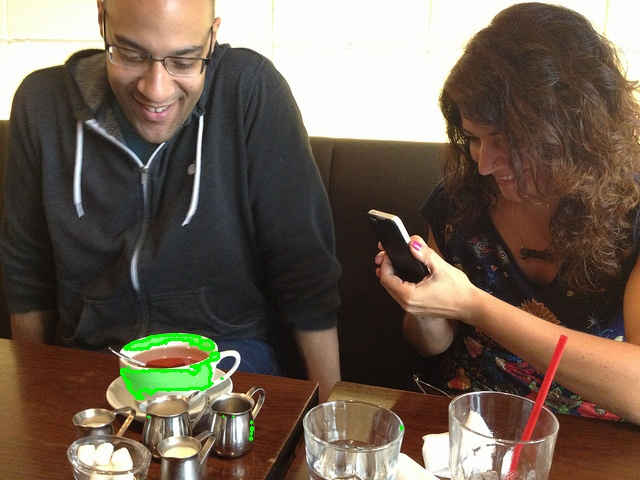}      \\[-0.25ex]
      (a) Image & 
      (c) GSVA \cite{xia2024gsva} &
      (d) PolyFormer \cite{liu2023polyformer} &
      (e) LISA \cite{lai2024lisa} &\\[0.75ex]
      \includegraphics[width=\www,height=\hhh]{imgs/paper-figs/overlays/134_mask.png}&
      \includegraphics[width=\www,height=\hhh]{imgs/paper-figs/qualitative-imgs4/masks/000000066771_masked_img_0_PWL-cropped.png}&
      \includegraphics[width=\www,height=\hhh]{imgs/paper-figs/qualitative-imgs4/masks/000000066771_masked_img_0_PWL+Mistral-cropped.png}&
      \includegraphics[width=\www,height=\hhh]{imgs/paper-figs/qualitative-imgs4/masks/000000066771_masked_img_0_GP-cropped.png}      \\[-0.25ex]
      (b) Ground truth &
      (f) \textbf{SANSA} (DISP w/o LLM) &
      (g) \textbf{SANSA} (DISP) & 
      (h) \textbf{SANSA} (EXSP) 
      \end{tabular}
      }
    \caption{\textbf{Qualitative segmentation results} for (a) input image and (b) object ground truth, described by a semantic-agnostic prompt from the HP test set:
    \texttt{\small "Segment the rectangular dark shape with rounded corners and grey, almost white reflective borders".}
    We compare (c)-(e) pretrained segmentation VLMs and (f)-(h) our finetuned SANSA models from LISA.
    Our SANSA models accurately segment the object, whereas the pretrained VLMs fail to interpret the semantic-agnostic description.
}
    \label{fig:qualitative}
\end{figure*}

\begin{figure}[htb]
      \centering
      {\footnotesize
      \begin{tabular}{@{\hspace{0mm}}cc@{\hspace{0mm}}}
      \includegraphics[height=0.38\columnwidth]{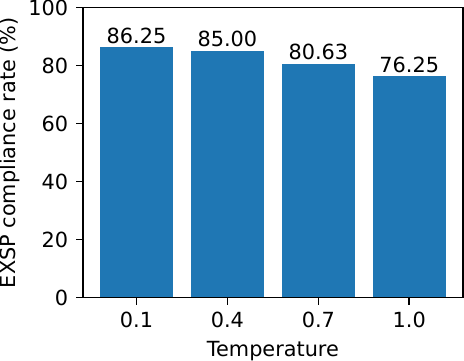}&
      \includegraphics[height=0.4\columnwidth]{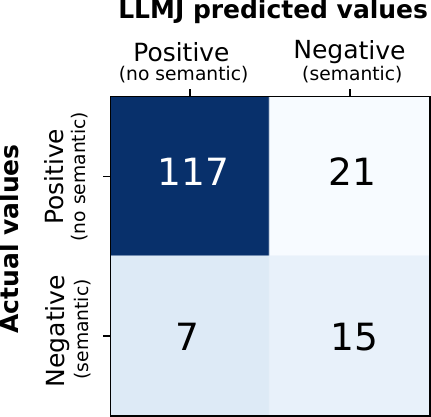} \\[0.25ex]
      (a)  
      Human evaluation over temperature &
      (b) LLMJ evaluation 
      \end{tabular}
      }
    \caption{
        \textbf{EXSP compliance with the SANSA constraint} (on 160 prompts).
        (a) Human evaluation of semantic-agnostic compliance over different temperatures for the description VLM.
        (b) Automatic evaluation of LLMJ predictions for a temperature of 0.1.   
        }
    \label{fig:exsp-llmj-robustness}
\end{figure}

\medskip\noindent\textbf{Segmentation Accuracy.}
Quantitative SANSA segmentation results are shown in Table~\ref{table:sansa-results}.
We evaluate both pretrained VLMs (GSVA, PolyFormer, and LISA) and models finetuned on our SANSA prompts (DISP w/o LLM, DISP, and EXSP). The evaluation is conducted across multiple test sets: DISP w/o LLM, DISP, EXSP, EXSP+LLMJ, the combined average, and the Human Prompt (HP) set.
{Note that the EXSP+LLMJ test set contains less images, as around 25\% were filtered by LLMJ.}

All models finetuned on SANSA prompts consistently and significantly outperform the pretrained models across all test sets.
For finetuned models, the highest performance is generally observed when training and test sets use the same type of prompts, which naturally reflects the matching data distributions.
For the DISP strategy, LLM post-processing improves performance (DISP w/o LLM vs DISP) on average, highlighting the benefit of producing well-formed prompts.
{On the filtered EXSP+LLMJ test set, we can see that both DISP models achieve slightly better results than on the EXSP test set, which is consistent with their training on a restricted dictionary without semantic.}
Finally, the EXSP strategy achieves the highest average mIoU across automatic test sets, as well as on the HP set.
This suggests that EXSP generates prompts that are particularly appropriate to the SANSA task and support generalization to human-written semantic-agnostic descriptions.

Importantly, in the supp. mat. we show that finetuning on SANSA prompts does not reduce performance on the original semantic segmentation task, remaining comparable to pretrained LISA.

\medskip\noindent\textbf{Qualitative Results.}
Fig.~\ref{fig:qualitative} compares all evaluated models on a SANSA segmentation example.
Our finetuned models accurately segment the object, whereas the pretrained models fail to interpret the semantic-agnostic description.
{More examples are given in the supp. mat.}

\medskip\noindent\textbf{Robustness of EXSP \& LLMJ.}
As EXSP can generate prompts that sometimes include unintended semantic information, we study the robustness of the EXSP strategy and of the LLMJ filtering.
First, we manually evaluated EXSP prompts on a subset of 160 images (distinct from the HP test set), marking whether each prompt contained semantic information.
This evaluation was performed for multiple temperatures of the description VLM to assess its influence on semantic leakage.
Fig.~\ref{fig:exsp-llmj-robustness}(a) shows that the proportion of semantic-agnostic prompts tends to decrease with increasing temperature.
We therefore retained the 
temperature of 0.1 in our experiments.

Next, we assess LLMJ filtering by comparing its automatic classification against manual annotations.
This allows to measure how accurately LLMJ detects prompts containing semantic information.
The resulting confusion matrix in Fig.~\ref{fig:exsp-llmj-robustness}(b) shows, on this subset of examples, an overall accuracy of 82.5\% and a high precision of 94\%, indicating very few false positives.

\section{Conclusion}

We introduced semantic-agnostic and shape-aware (SANSA) segmentation, a vision–language paradigm that removes semantic object categories from prompts and relies on visual properties such as shape, geometry, color, and texture.
This formulation exposes a key limitation of existing
segmentation VLMs, which primarily rely on semantic alignment rather than intrinsic visual reasoning.

We proposed an automated pipeline for SANSA prompt generation, based on two complementary strategies, enabling scalable construction of segmentation-agnostic segmentation datasets. 
We have shown that finetuning a state-of-the-art VLM on these prompts leads to substantial improvements on this new SANSA task, both quantitatively and qualitatively.
SANSA provides a promising direction for extending VLM 
beyond semantic-only understanding, particularly for 
domains where semantic labels are unavailable or insufficient.

\label{sec:conclusion}

\FloatBarrier
\clearpage

\FloatBarrier
\clearpage

\makeatletter
\let\maketitle\savedmaketitle
\let\@maketitle\saved@maketitle
\makeatother

\SetPaperTitleSupMat
\maketitle

\setcounter{section}{0}
\setcounter{figure}{0}
\setcounter{table}{0}
\setcounter{equation}{0}

\renewcommand{\thesection}{S\arabic{section}}
\renewcommand{\thefigure}{S\arabic{figure}}
\renewcommand{\thetable}{S\arabic{table}}
\renewcommand{\theequation}{S\arabic{equation}}

\maketitle

\setcounter{section}{0}
\renewcommand{\thesection}{\Alph{section}}
\renewcommand{\thesubsection}{\thesection.\arabic{subsection}}

This supplementary material provides additional details about our two proposed SANSA prompting strategies: (i) Dictionary-based segmentation prompts (DISP) and (ii) Example-based segmentation prompts (EXSP).
We also include an experimental baseline on standard semantic segmentation to demonstrate that finetuning on SANSA prompts does not compromise performance on the original task.
Finally, we include additional qualitative segmentation results on semantic-agnostic prompts.

\section{DISP Strategy Details}

\subsection{DISP Dictionary}

As explained in Sec. 2.1, the DISP strategy is based on a predefined dictionary of semantic-agnostic words, restricting the description VLM to a constrained vocabulary.
These words are organized into semantic-agnostic attributes such as colors, textures, shapes, patterns, lighting and connector words.
The complete list of words grouped by attribute is as follows:

\begin{itemize}[leftmargin=1em]
\item \textbf{Colors}: \textit{amber, beige, black, blue, bronze, brown, burgundy, copper, coral, cream, cyan, dark, gold, gray, green, grey, indigo, ivory, khaki, light, lime, magenta, maroon, multicolored, mustard, navy, olive, orange, peach, pink, purple, red, salmon, silver, tan, teal, turquoise, violet, white, yellow.}
\item \textbf{Textures}: \textit{bumpy, coarse, fabric, fabric-like, fuzzy, glassy, glossy, grainy, leathery, metallic, metallic-like, opaque, paper-like, plastic-like, polished, porous, rough, rubber-like, shiny, silky, smooth, soft, stone-like, translucent, transparent, velvety, wood-like, wooden, woolly, wrinkled.}
\item \textbf{Shapes}: \textit{angular, arched, asymmetrical, bulky, circular, clustered, conical, contour, corners, curved, cylindrical, domed, edges, elongated, flat, flattened, form, hexagonal, irregular, layered, narrow, outline, oval, pointed, polygonal, profile, proportions, rectangular, rounded, sharp, short, silhouette, slender, spherical, square, stacked, straight, surface, symmetrical, tall, tapered, triangular, wide.}
\item \textbf{Patterns}: \textit{blended, checkered, dotted, faded, gradient, grid, irregular, lattice, marbled, pattern, plaid, repeated, spotted, striped, swirled, textured, uniform, zigzag.
\item \textbf{Lighting}: bright, contrasted, dim, faint, glossy, harsh, highlighted, matte, muted, radiant, reflective, shadowed, soft, strong, subtle.}
\item \textbf{Connectors}: \textit{,, -, ., ;, a, across, adjacent, along, although, an, and, appears, area, around, as, at, between, bottom, center, contains, contrast, darkened, deep, displays, entire, entirely, features, from, has, highly, in, lightened, middle, mostly, near, of, on, overall, pale, part, partly, presents, region, resembles, rich, section, seems, shows, side, slightly, somewhat, subtle, surface, that, the, throughout, to, top, towards, very, where, which, while, whole, with.}
\end{itemize}

\subsection{DISP Instruction Prompt}

Once the description VLM has been constrained by the predefined dictionary, the DISP instruction prompt is as follows:

\begin{small}
\parbox[t]{8cm}{\texttt{"Ignore any black/empty background. Describe ONLY the visible object using strictly visual cues. Use low-level attributes: colors/tones, surface qualities, overall outline/proportions and simple patterns."}}
\end{small}

\subsection{DISP post-processing with LLM}

Object descriptions obtained by the DISP strategy are reformulated by an external LLM with the following prompt:

\begin{small}
\parbox[t]{8cm}{\texttt{"Instruction: Convert the input into a single, concise segmentation command. Start with 'Segment ...' and use only the given words."}}
\end{small} \\[0.1em]

Some examples of this LLM post-processing are shown in Table~\ref{table:LLM-refinement}.

\begin{table}[h]
    \caption{Examples of LLM post-processing for DISP object descriptions.}
    \vspace{0.2cm}
    \renewcommand{\arraystretch}{1}
    \newcommand{\sepp}{3mm}
    \centering
    {\scriptsize
        \begin{tabular}{
            @{\hspace{0mm}}p{4.1cm}@{\hspace{\sepp}}
            p{3.6cm}@{\hspace{0mm}}
        }
        \multicolumn{1}{@{\hspace{0mm}}c@{\hspace{0mm}}}{\textbf{DISP w/o LLM}}
        & \multicolumn{1}{@{\hspace{0mm}}c@{\hspace{0mm}}}{\textbf{DISP}} \\
        \toprule
        \parbox[t]{4.1cm}{
            - cylindrical in form\\
            - dark brown with a glossy surface\\
            - smooth and reflective\\
            - top appears slightly rounded\\
            - overall proportions elongated, tapered towards the bottom\\
            - subtle light and shadow contrast, with a light area near the top center\\
            - overall dark with a gradient of light towards the center\\
            - surface appears to be polished and shiny\\
            - edges soft and rounded, with a subtle gradient of light and dark\\
            - overall silhouette elongated and tapered, with a rounded top and bottom\\
            - light area near the top center, with a gradient
        }
        & \parbox[t]{3.6cm}{Segment the object as a dark brown, cylindrical, elongated, and tapered structure with a slightly rounded top and bottom. It has a polished, glossy, and reflective surface with soft, rounded edges. The object exhibits a subtle light and shadow contrast, with a light area} \\
        \midrule
        \parbox[t]{4.1cm}{
            - green, textured surface\\
            - irregular, somewhat rounded outline\\
            - rough, bumpy surface\\
            - overall oval-like in outline\\
            - subtle, indigo and brown shadow around the edges\\
            - slightly fuzzy, soft edges\\
            - appears to be a clustered, conical form\\
            - light and dark green gradient\\
            - surface has a slightly glossy, reflective area in the center\\
            - overall, the surface has a somewhat irregular, asymmetrical pattern. \\
            - green, textured surface\\
            - irregular, somewhat rounded outline\\
            - rough, bumpy surface\\
            - overall
        }
        & \parbox[t]{3.6cm}{
            Segment irregular, oval-shaped, textured and bumpy green object with a rough, somewhat rounded outline. It has a subtle indigo and brown shadow around the edges, fuzzy and soft edges, and a clustered, conical form. The surface exhibits a light and dark green
        } \\
        \bottomrule
        \end{tabular}
        }
    \label{table:LLM-refinement}
\end{table}

\section{EXSP Strategy Details}

\subsection{EXSP Instruction Prompt}

As explained in Sec. 2.1, the EXSP strategy builds upon an instruction prompt, which combines positive and negative examples.
To design this prompt, we first conducted experiments with several alternative instruction prompt formulations, each with different levels of guidance (from IP1 to IP5), shown in Fig.~\ref{fig:EXSP-prompt}.
IP5 corresponds to the EXSP instruction prompt used in the experiments of the main paper.

\subsection{Semantic-Agnostic Descriptions Challenge} 

We evaluated the robustness of these instruction prompts across multiple description VLMs, specifically: BLIP2 (OPT-2.7B) \cite{li2023blip}, MiniCPM-V 2~\cite{yao2024minicpmv}, LLaVa 1.6 (Mistral-7B) \cite{liu2023improved}, and InternVL 2.5 \cite{chen2024expanding}.

Table~\ref{table:vlm-choice} shows the description prompts generated by each description VLM and each instruction prompt.
The results clearly show, for this example image, that most configurations of VLM and IP tend to leak semantic information into the object description even when asked not to.
Only the combination of InternVL 2.5 and IP5 produces a proper semantic-agnostic description for this case and has given the best overall results in our experiments.

Table~\ref{table:cropping-strategies} shows that cropping strategies may also impact the description VLM quality. 
Using only cropped objects within their bounding box improves the VLM's ability to produce semantic-agnostic descriptions and prevents information from nearby objects from being included. We adopt this cropping strategy in our work.

\section{LLMJ Filtering for EXSP}

As shown in previous experiments,
even after careful guidance, the EXSP strategy may still produce prompts containing unintended semantic information.
Therefore, the description VLM (InternVL 2.5~\cite{chen2024expanding}) can be employed as LLM-as-a-judge (LLMJ), using text only, to classify each EXSP description as either semantic or semantic-agnostic.
The full prompt for this LLMJ filtering is shown in Fig.~\ref{fig:LLMJ-prompt}.

\section{Semantic Segmentation Baseline}

In addition to evaluating our models on the SANSA task, we assess their robustness on the original semantic segmentation task.
This allows us to verify that finetuning on semantic-agnostic prompts does not compromise their ability to segment objects when semantic information is explicitly provided.
For this evaluation, we use the same COCO test subset introduced in Sec. 3.1, and directly instruct the model to segment each object by mentioning its class.
Specifically, for each object, we use the following segmentation prompt:

\parbox[t]{8cm}{\texttt{"Can you segment the \{obj\_category\} in this image?"}}. 
\vspace{-0.1em}

Table~\ref{table:ssd-results} shows that the performance of our finetuned models remains very comparable to the pretrained LISA model, with less than 1\% drop in mIoU, indicating that finetuning on SANSA prompts does not meaningfully reduce their capabilities on standard semantic segmentation.

\begin{table}
    \caption{Comparison between our finetuned SANSA models and SOTA segmentation VLMs on the semantic segmentation baseline, in terms of mIoU (\%). 
    Finetuning on SANSA prompts does not degrade performance on the standard semantic segmentation task.}
    \resizebox{\linewidth}{!}{
    \vspace{0.1cm}
    \renewcommand{\arraystretch}{1}
    \newcommand{\sepp}{2mm}
    \centering
    {\small
        \begin{tabular}{
            c@{\hspace{\sepp}}
            c@{\hspace{\sepp}}
            c@{\hspace{\sepp}}
            c@{\hspace{\sepp}} | 
            c@{\hspace{\sepp}}
            c@{\hspace{\sepp}}
            c@{\hspace{\sepp}}
        }
 \\
        \text{LAVT \cite{yang2022lavt}}
        & \text{GSVA 7B \cite{xia2024gsva}}
        & \text{PolyFormer \cite{liu2023polyformer}}
        & \text{LISA 7B \cite{lai2024lisa}}
        & \text{DISP w/o LLM}
        & \text{DISP}
        & \text{EXSP}
        \\
        \toprule
        25.14 & 25.23 & 39.46 & \textbf{45.40} & 44.95 & 45.17 & \underline{45.27} \\
        \bottomrule
        \end{tabular}
        }
    }
    \label{table:ssd-results}
\end{table}

\section{Additional qualitative results}

Additional qualitative results are shown in Fig.~\ref{fig:additional-qualitative}.

\begin{figure*}
\begin{tiny}
\begin{tabular}{p{0.01\textwidth}@{\hspace{2mm}}p{0.97\textwidth}}
\\
     \textbf{IP1}: &\texttt{"Describe the object."}\\[1ex]
     \textbf{IP2}: & \texttt{"You are a helpful and precise low-level visual assistant. Describe the object."}\\[1ex]
     \textbf{IP3}: & \texttt{"You are a helpful and precise low-level visual assistant. Your goal is to describe the object in the image without using any high-level semantic or conceptual information."} \\[4ex]
     \textbf{IP4}: &\texttt{"You are a helpful and precise low-level visual assistant. Your goal is to describe the object in the image without using any high-level semantic or conceptual information. To describe the object, you can only use geometric, shape, and other visual properties, such as: shape, geometry, texture, surface patterns, materials, colors, brightness, dimensions. You can choose between these properties, or add additional ones, according to the most dominant ones in the images but never name the object or subparts of the objet in any way. Any semantic information or any knowledge about what the object is or its function is prohibited. You cannot name the object class and any of its parts/components. Do not describe the background."}\\[5ex]
     
     \textbf{IP5}: &\texttt{"You are a helpful and precise low-level visual assistant. Your goal is to describe the object in the image without using any high-level semantic or conceptual information. To describe the object, you can only use geometric, shape, and other visual properties, such as: shape, geometry, texture, surface patterns, materials, colors, brightness, dimensions. You can choose between these properties, or add additional ones, according to the most dominant ones in the images but never name the object or subparts of the objet in any way. Any semantic information or any knowledge about what the object is or its function is prohibited. You cannot name the object class and any of its parts/components. Do not describe the background.}\\
     &\texttt{Here are some examples of expected words to use (RIGHT)/not to use (WRONG), problematical words are written in capital letters: }\\
     &\texttt{- Tomato:}\\
     &\texttt{WRONG: 'The object is a TOMATO...'}\\
     &\texttt{WRONG: 'The object is a VEGETABLE that CAN BE EATEN...' }\\
     &\texttt{RIGHT: 'The object is a circular and red, with a smooth surface and a green pattern on top. It has a green rectangular subpart on top. The surface appears to be highly reflective. The overall shape of the object is consistent and symmetrical.' }\\
     &\texttt{- Green mug:}\\
     &\texttt{WRONG: 'The object is a MUG...' }\\
     &\texttt{WRONG: 'The object CAN BE USED TO DRINK COFFEE OR TEA...' }\\
     &\texttt{RIGHT: 'The object is cylindrical with an open top and a curved part, with a smooth green texture and is asymmetrical.'}\\
     &\texttt{- Bike:}\\
     &\texttt{WRONG: 'The object has WHEELS and PEDALS...' }\\
     &\texttt{WRONG: 'The object CAN BE USED BY A PERSON TO MOVE...' }\\
     &\texttt{RIGHT: 'The object is a symmetric and complex arrangement of multiple elongated tubular shapes, with metallic texture. The object also has two large circular parts on each side, with repeated straight lines inside arranged in a radial pattern.' }\\
     &\texttt{- Person: }\\
     &\texttt{WRONG: 'The object has a white SHIRT and a blue SHORT...' }\\
     &\texttt{WRONG: 'The PERSON IS WEARING...' }\\
     &\texttt{RIGHT: 'The object is white and blue, elongated, cylindrical. It has a yellow and white pattern on the top.'"} \\[-1ex]
\end{tabular}
\end{tiny}
\caption{Instruction prompts (IP1 to IP5) tested for SANSA segmentation prompt generation with the EXSP strategy. IP5 is the final EXSP instruction prompt used in the experiments of the main paper.}
\label{fig:EXSP-prompt}
\end{figure*}

\begin{figure*}
\begin{tiny}
\begin{tabular}{p{\textwidth}}
\\
\texttt{"TESTED\_SENTENCE: '\{response\}' Does TESTED\_SENTENCE contain any semantic information, 'YES' or 'NO' ? }\\
\texttt{For example, you have to answer 'YES' for the following sentences:  }\\
\texttt{(1): 'The object is a clock, with black boundaries and branches.'}\\
\texttt{(2): 'The image shows a grey or silver airplane flying in the sky. The airplane has multiple rivets on the sides.'}\\
\texttt{(3): 'The object is an animal, black and brown, fluffy texture. Its claws are big and the dog is opening his mouth.'}\\
\texttt{(4): 'The image shows a person, oval shape, white and blue.'}\\
\texttt{(5): 'The object has a golden-brown crust. The surface has a glossy appearance, likely indicating a baked or toasted material.'}\\
\texttt{(6): 'The object in the image is a large aircraft, specifically a Royal Air Force (RAF) aircraft.'}\\
\texttt{(7): 'The object is oval, green and yellow, with a smooth texture, without pattern, likely made of vegetables, with slight reflection and is medium-sized.'}\\
\texttt{(8): 'The object is oval, red and white, with a smooth texture, without pattern, likely made of dough or pastry, with slight reflection.'}\\
\texttt{(9): 'The object is rectangular, red and white, with a smooth texture, without pattern, likely made of dough or pastry, with slight reflection.'}\\
\texttt{(10): 'The image shows a person. The person is wearing a black shirt. The person has a white beard. The person is smiling.'}\\
\texttt{Describing the object, sub-object, its class or sub-class are semantic information contrary to shape, geometry, texture, surface patterns, materials, colors, brightness, shadows, reflections, and any measurable or mathematical properties (curvature, dimensions) information that you have NOT to consider as semantic information.}\\
\texttt{For example, you have to answer 'NO' for the following sentences:}\\
\texttt{(1): 'The object is circular, red and white, with a smooth texture, without pattern, likely metal or plastic, with slight reflection and is large.'}\\
\texttt{(2): 'The image shows a grey or silver cylindrical object. The object is elongated and the texture is like metal.'}\\
\texttt{(3): 'The object is oval, black and brown, fluffy texture. No reflection.'}\\
\texttt{(4): 'The object is oval, white, smooth texture, without pattern, likely ceramic or porcelain, with slight reflection and is large.'}\\
\texttt{(5): 'The object is rectangular, purple, with a smooth texture, without pattern, likely metal or plastic, with slight reflection and is large.'}\\
\texttt{(6): 'The object is rectangular, brown and yellow, with a smooth texture, has a striped pattern, likely made of leather or fabric, with slight reflection and is medium-sized.'}\\
\texttt{(7): 'The object is rectangular, with a surface pattern of green, red, and yellow colors. The texture appears to be rough, and there is a slight reflection. The object is medium-sized.'}\\
\texttt{(8): 'The object is elongated, white, with a fluffy texture, without pattern, likely made of wool or fur, with slight reflection and is medium-sized.'}\\
\texttt{(9): 'The object is oval, yellow and green, with a smooth texture, without pattern, likely made of plastic or metal, with slight reflection.'}\\
\texttt{(10): '(1): 'The object is elongated, silver and grey, with a smooth texture, without pattern, likely metal, with slight reflection.''}\\
\texttt{(11): 'The object is rectangular, brown, with a smooth texture, has a striped pattern, likely made of leather or fabric, with slight reflection.'}\\
\texttt{(12): 'The object is cylindrical, blue, with a smooth texture, decorated with white star patterns, likely metal or plastic, with slight reflection.'}\\
\texttt{(13): 'The object is circular, white and green, with a smooth texture, without pattern, likely ceramic or porcelain, with slight reflection.'}\\
\texttt{(14): 'The object is cylindrical, transparent, with a smooth texture, without pattern, likely glass, with slight reflection.'}\\
\texttt{(15): 'The object is elongated, blue and grey, with a smooth texture, without pattern, likely metal, with slight reflection.'}\\
\texttt{(16): 'The object is irregularly shaped, brown and black, with a fluffy texture. No reflection.'}\\
\texttt{(17): 'The object is cylindrical, transparent, with a smooth texture, without pattern, likely made of plastic, with slight reflection.'}\\
\texttt{(18): 'The object is elongated, yellow, with a smooth texture, without pattern, likely made of a soft material, with slight reflection.'"}
\end{tabular}
\end{tiny}
\caption{Prompt for LLMJ filtering on EXSP. \parbox[t]{2cm}{\texttt{\{response\}}} indicates the input object description.}
\label{fig:LLMJ-prompt}
\end{figure*}

\begin{table*}[]
    \caption{Comparison of several description VLMs across instructions prompts IP1 to IP5 for the EXSP strategy. Red and green background colors respectively indicate failure and success in respecting the semantic-agnostic constraint.
    For this example image, the results show that ensuring strict adherence to this constraint remains challenging for most VLM and IP configurations.
    }
    \resizebox{\textwidth}{!}{
    \vspace{0.2cm}
    \renewcommand{\arraystretch}{1.3}
    \newcommand{\sepp}{2mm}
    \centering
    {\scriptsize
        \begin{tabular}{
            l@{\hspace{3mm}}
            p{3cm}@{\hspace{\sepp}}
            p{3cm}@{\hspace{\sepp}}
            p{3cm}@{\hspace{\sepp}}
            p{3cm}@{\hspace{\sepp}}
            p{3cm}@{\hspace{\sepp}}
        }
        \\
        & \multicolumn{5}{c}{
            \includegraphics[height=2.5cm]{imgs/paper-figs/mask-and-crop-figs/autre/96549/000000096549_ann156368_og.png}
        } \\
        \centering
        & \multicolumn{1}{c}{\textbf{IP1}}
        & \multicolumn{1}{c}{\textbf{IP2}}
        & \multicolumn{1}{c}{\textbf{IP3}}
        & \multicolumn{1}{c}{\textbf{IP4}}
        & \multicolumn{1}{c}{\textbf{IP5}} \\
        \toprule
        \multirow{2}{*}{BLIP2 \cite{li2023blip}} & \cellcolor{myred} egyptair boeing 767-300
        & \cellcolor{myred} egyptian airways boeing 777-300
        & \cellcolor{myred} The object is an egyptian airplane
        & \cellcolor{myred} 
        & \cellcolor{myred} The object is a person wearing a white shirt and blue shorts.
        \\
        \midrule
        \multirow{9}{*}{MiniCPM-V 2 \cite{yao2024minicpmv}} & 
        \cellcolor{myred} The image prominently features a large airplane, predominantly white with blue and black accents. It's on the ground at an airport runway during what appears to be dusk or evening hours due to low light conditions in some parts of the scene.
        & \cellcolor{myred} This is an airplane, specifically a large passenger jet with the word "Egypt" written on it. 
        & \cellcolor{myred} The image features a large airplane with blue and white colors, prominently displaying the word "Egypt" on its side. The aircraft is situated at an airport runway during dusk or night time as indicated by the dark sky background.
        & \cellcolor{myred} airplane, wing
        & \cellcolor{myred} The object is a large airplane with blue and white colors. It has multiple windows on its side, some of which are open or closed depending upon the angle from where it's being viewed in this image. The plane appears to be stationary as if waiting for takeoff at an airport runway during dusk hours due to low light conditions.
        \\
        \midrule
        \multirow{15}{*}{LLaVa 1.6 \cite{liu2023improved}} & 
        \cellcolor{myred} The image shows a large commercial airplane on the runway at an airport. The airplane is predominantly white with a blue and white logo on the tail, which is characteristic of the Egyptian national airline, EgyptAir. The registration number "YA 264" is visible on the side of the aircraft. The airplane is equipped with two engines, one on each wing, and the landing gear is down, indicating that it is either preparing for takeoff or has recently landed. The sky is dark, suggesting that the photo was taken either during the evening or at night.
        & 
        \cellcolor{myred} The object in the image is an airplane. It is a large commercial aircraft with a predominantly white body and a tail fin that features a design with blue and black colors, which is characteristic of the livery of EgyptAir. The airplane is on the ground, likely at an airport, as indicated by the runway and the airport infrastructure in the background. The aircraft is equipped with two engines, one on each wing, and the landing gear is down, suggesting that it is either preparing for takeoff or has recently landed. The sky is dark, indicating that the photo was taken either
        & 
        \cellcolor{myred} The image shows a large commercial airplane on a runway at an airport. The airplane is predominantly white with a blue and black design on its tail and body. The registration number "YA 264" is visible on the side of the aircraft. The airplane is equipped with two engines, one on each wing. The runway has markings and lights, indicating it is an active runway. The sky is dark, suggesting it might be evening or nighttime. There are no visible people or vehicles in the immediate vicinity of the airplane.
        & \cellcolor{myred} The image features a large, elongated object with a predominantly white body and a darker blue tail. The object has a series of windows aligned along its side, and there are visible engines under the wings. The object is positioned on a runway, and there are various markings and lights on the ground around it. The object appears to be stationary, and there is a clear sky in the background. 
        & \cellcolor{myred} The object is a large, elongated, cylindrical shape with a white and blue color scheme. It has a blue and white pattern on the top. The object has two large circular parts on each side, with repeated straight lines inside arranged in a radial pattern. The object also has two smaller circular parts on the bottom. The object is highly reflective, with a metallic texture. The object is asymmetrical. The object is on a runway. The object has two large circular parts on the bottom. The object is highly reflective, with a metallic texture. The
        \\
        \midrule
        \multirow{17}{*}{InternVL 2.5 \cite{chen2024expanding}} & 
        \cellcolor{myred} The object in the image is a large commercial airplane. It is a wide-body jet, likely an Airbus A330, based on its size and shape. The aircraft is painted in the livery of EgyptAir, featuring a white fuselage with a teal and black design on the tail and engines. The plane is on the ground, possibly taxiing or preparing for takeoff or landing. The setting appears to be an airport runway during twilight or early evening, as indicated by the lighting and the darkening sky.
        & \cellcolor{myred} The object in the image is a large commercial airplane on a runway. The aircraft is predominantly white with blue and teal accents. The word "EGYPTAIR" is prominently displayed on the fuselage, indicating that it belongs to the Egyptian airline. The plane is a wide-body jet, likely an Airbus A330, based on its size and design features such as the twin engines and the shape of the wings. The aircraft is positioned on the ground, with its landing gear visible and the runway lights illuminated in the background. The setting appears to be either dawn or d
        & \cellcolor{myred} The image shows a large commercial airplane on a runway. The aircraft has a predominantly white fuselage with blue and teal accents and markings. The word "EGYPTAIR" is written in large letters along the side of the plane. The tail features a teal and blue design with a logo. The airplane is equipped with two large engines, one under each wing. The landing gear is visible, and the plane appears to be either taking off or landing. The background shows a cloudy sky and some airport infrastructure.
        & \cellcolor{myred} The image shows a large aircraft with a predominantly white body. The aircraft has a streamlined fuselage with a smooth, curved surface. The wings are wide and extend horizontally from the fuselage, with a noticeable upward angle at the tips. The tail section features a vertical stabilizer with a teal-colored design. The engines are mounted under the wings, with a circular shape and a metallic finish. The landing gear is visible, with multiple wheels supporting the aircraft. The overall color scheme includes white, teal, and black accents. The lighting suggests it is either dawn or d
        & \cellcolor{mygreen} The object is a large, elongated, cylindrical shape with a smooth white surface. It has a blue and teal pattern running along its length, with a teal tail fin. The surface appears to be highly reflective, and the overall shape is consistent and symmetrical. The object has two large circular parts on each side, with repeated straight lines inside arranged in a radial pattern
        \\
        \bottomrule
        \end{tabular}
        }
    \label{table:vlm-choice}
    }
\end{table*}

\begin{table*}[]
    \caption{Comparison of a description VLM InternVL 2.5 \cite{chen2024expanding} across instruction prompts IP1 to IP5 for the EXSP strategy, using different image pre-processing methods (cropping and object masking).
    Red and green background colors respectively indicate failure and success in respecting the semantic-agnostic constraint.
    Cropping and object masking help the VLM to focus on the object to describe.
    }
    \resizebox{\textwidth}{!}{
    \vspace{0.3cm}
    \renewcommand{\arraystretch}{1.3}
    \newcommand{\sepp}{2mm}
    \centering
    {\scriptsize
        \begin{tabular}{
            l@{\hspace{5mm}}
            p{3.5cm}@{\hspace{2mm}}
            p{3.5cm}@{\hspace{2mm}}
            p{3.5cm}@{\hspace{2mm}}
            p{3.5cm}@{\hspace{2mm}}
        }
        \\
        & \multicolumn{1}{c}{\includegraphics[height=2.5cm]{imgs/paper-figs/mask-and-crop-figs/bananas/12667/000000012667_ann1043443_og.png}}
        & \multicolumn{1}{c}{\includegraphics[height=2.5cm]{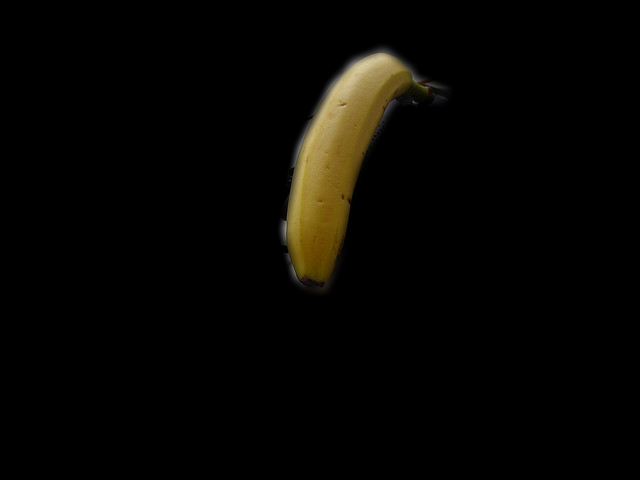}}
        & \multicolumn{1}{c}{\includegraphics[height=2.5cm]{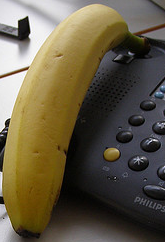}}
        & \multicolumn{1}{c}{\includegraphics[height=2.5cm]{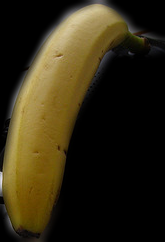}}
        \\
        \toprule
        \multirow{9}{*}{\textbf{IP1}} & 
        \cellcolor{myred} The image shows a banana placed on top of a black Philips telephone. The banana is positioned in such a way that it appears to be part of the phone's handset, with its curved shape mimicking the phone's design. The telephone is situated on a desk, and there is a notepad with handwritten notes visible in the background. &
        \cellcolor{myred} The object in the image is a banana. It appears to be a single, ripe banana with a yellow peel. The background is completely black, which makes the banana stand out prominently. The banana is positioned in the center of the image, slightly tilted, giving it a dynamic appearance. The lighting highlights the curve of the banana, emphasizing its natural shape and texture. &
        \cellcolor{myred} The object in the image is a banana. It is a long, yellow fruit with a curved shape and a green stem at one end. The banana is placed on top of a black telephone, which has several buttons and a speaker grille. &
        \cellcolor{myred} The object in the image is a banana. It appears to be ripe, with a yellow peel and some brown spots. The banana is slightly curved and is positioned against a dark background. 
        \\
        \midrule
        \multirow{9}{*}{\textbf{IP2}} &
        \cellcolor{myred} The image shows a black Philips cordless telephone placed on a desk. A ripe banana is resting on the handset of the phone, with its stem pointing towards the base of the phone. The desk also has a stapler, some cables, and a notepad with handwritten notes. &
        \cellcolor{myred} The object in the image is a banana. It appears to be a single, ripe banana with a yellow peel. The background is completely black, which makes the banana stand out prominently. The banana is positioned in the center of the image and is slightly curved, typical of its natural shape. &
        \cellcolor{myred} The image shows a yellow banana resting on a black Philips telephone. The banana is positioned diagonally across the phone, with its stem end pointing towards the top right corner of the image. The phone has several buttons, including a yellow button and a star-shaped button, among others. The background appears to be a light-colored surface, possibly a table or desk. &
        \cellcolor{myred} The object in the image is a banana. It is a long, curved fruit with a yellow peel. The banana appears to be ripe, as indicated by its color and the presence of some brown spots on the peel. The background is dark, which makes the banana stand out prominently. \\
        \midrule
        \multirow{9}{*}{\textbf{IP3}} &
        \cellcolor{myred} The image shows a black landline telephone with a coiled cord on a white desk. A ripe yellow banana is placed on the handset of the phone. In the background, there is a stapler and some cables. To the right of the phone, there is a piece of paper with handwritten text. &
        \cellcolor{myred} The image shows a single banana with a yellow peel. The banana is positioned against a black background, giving it a prominent appearance. The peel has a smooth texture with some minor blemishes and a slight curve, indicating its natural shape. The banana is oriented vertically, with the stem end at the top and the curved end at the bottom. &
        \cellcolor{myred} The image shows a yellow banana with a green stem resting on a black electronic device. The device has several buttons, including a yellow button and a star-shaped button. The brand name "PHILIPS" is visible on the device. The background appears to be a white surface with some indistinct objects. &
        \cellcolor{myred} The image shows a yellow banana with a curved shape. It has a smooth surface with a few small brown spots. The banana is positioned against a dark background, and its stem is visible at the top. 
        \\
        \midrule
        \multirow{12}{*}{\textbf{IP4}} &
        \cellcolor{myred} The image features a yellow, elongated object with a curved surface and a pointed tip, resting on a black, rectangular device with a grid of buttons. The yellow object has a smooth texture and is positioned horizontally. The black device has a glossy surface and is rectangular in shape, with a screen displaying text and symbols. The yellow object is partially wrapped around the black device, with its tip pointing upwards. The black device has a coiled, black cord extending from one side. &
        \cellcolor{myred} The image features a curved, elongated object with a smooth, yellow surface. The object has a consistent color and texture throughout, with a slightly glossy appearance. The curvature is gentle, and the object appears to be uniform in thickness. The background is completely black, providing a stark contrast to the yellow object. &
        \cellcolor{myred} The image features a yellow, curved object with a smooth texture and a slightly pointed tip. It has a consistent color and appears to be resting on a flat, dark surface with a pattern of small holes. The object's surface is glossy, reflecting some light. &
        \cellcolor{mygreen} The image features a yellow, elongated object with a smooth, slightly curved surface. It has a consistent color and texture throughout, with a few minor imperfections and small dark spots. The object appears to be cylindrical in shape with a tapered end.
        \\
        \midrule
        \multirow{6}{*}{\textbf{IP5}} & 
        \cellcolor{mygreen} The object is a curved and elongated shape with a smooth yellow surface and a green stem-like protrusion on top. The surface appears to be slightly textured. The overall shape of the object is consistent and symmetrical. &
        \cellcolor{mygreen} The object is a curved and elongated shape with a smooth yellow surface. It has a consistent and symmetrical shape with a slightly darker yellow pattern along its length. The surface appears to be slightly reflective. &
        \cellcolor{mygreen} The object is a long and yellow, with a smooth surface and a green subpart on top. The surface appears to be slightly curved. The overall shape of the object is consistent and symmetrical. & 
        \cellcolor{mygreen} The object is a long, yellow, curved shape with a smooth surface. It has a greenish-brown tip at one end. The overall shape is consistent and symmetrical. The texture appears to be slightly bumpy with small indentations. The object is brightly colored and has a glossy appearance.
        \\
        \bottomrule
        \end{tabular}
        }
    \label{table:cropping-strategies}
    }
\end{table*}

\renewcommand{\www}{0.22\textwidth}
\renewcommand{\hhh}{0.13\textwidth}
\begin{figure*}[t]
      \centering
      {\footnotesize
      \begin{tabular}{@{\hspace{0mm}}c@{\hspace{4mm}}c@{\hspace{2mm}}c@{\hspace{2mm}}c@{\hspace{0mm}}}
      \includegraphics[width=\www,height=\hhh]{imgs/paper-figs/qualitative-imgs5/valise-og.png}&
      \includegraphics[width=\www,height=\hhh]{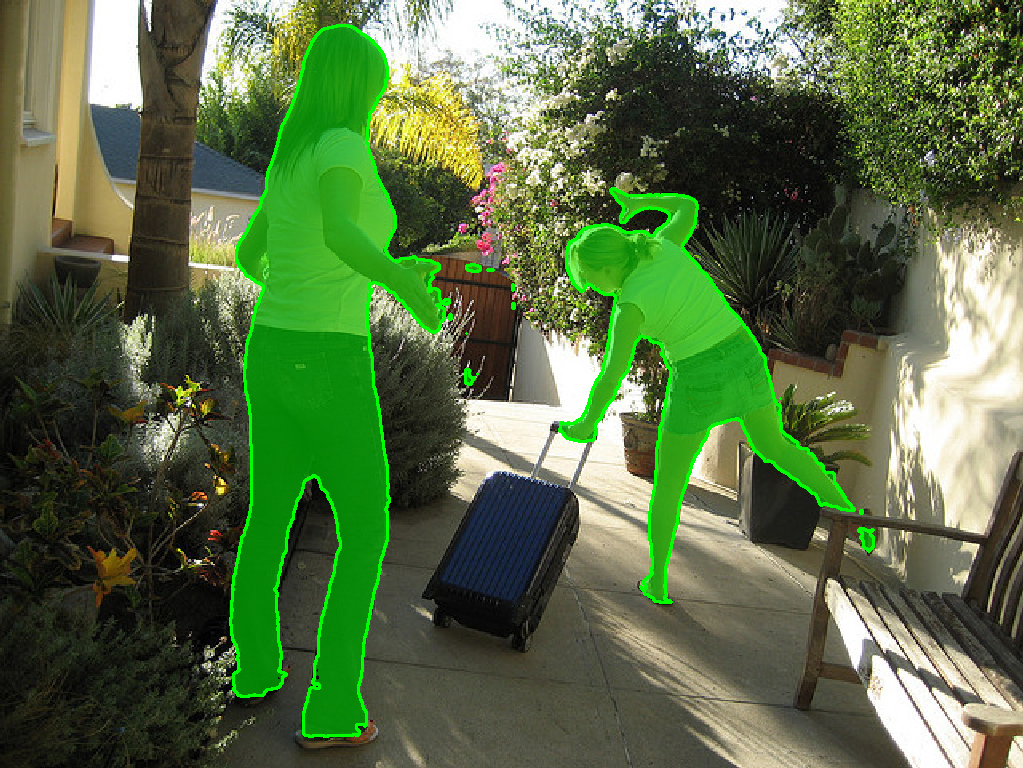}&
      \includegraphics[width=\www,height=\hhh]{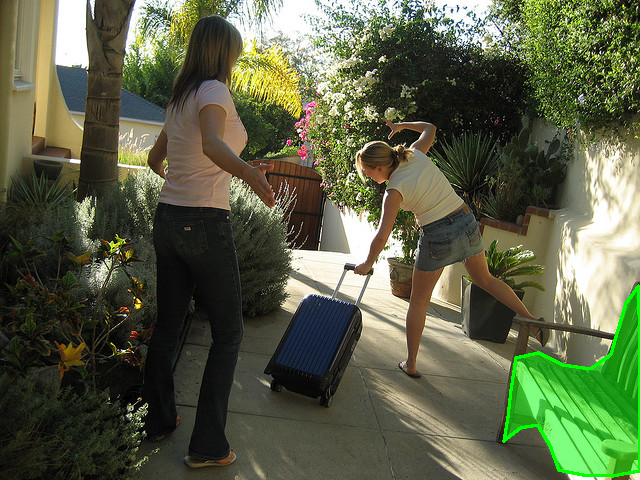}&
      \includegraphics[width=\www,height=\hhh]{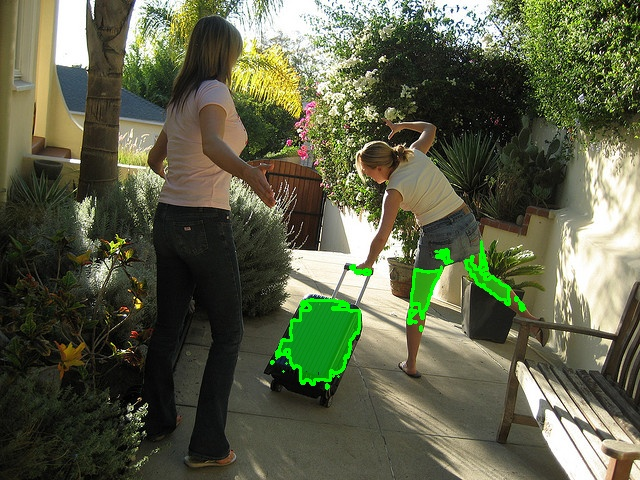}      \\[-0.25ex]
      (a) Image & 
      (c) GSVA \cite{xia2024gsva} &
      (d) PolyFormer \cite{liu2023polyformer} &
      (e) LISA \cite{lai2024lisa} \\[0.75ex]
      \includegraphics[width=\www,height=\hhh]{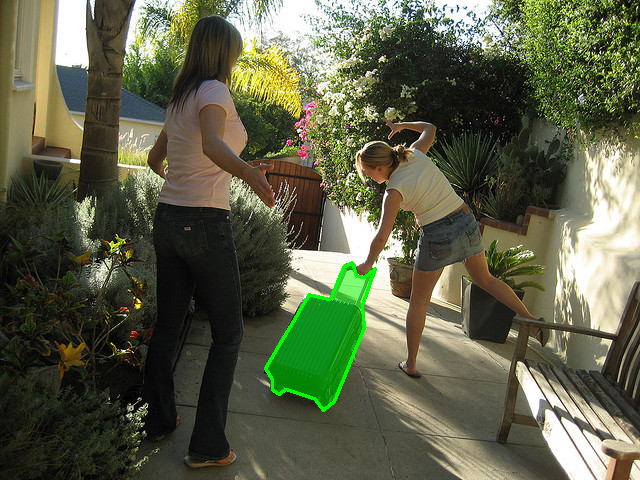}&
      \includegraphics[width=\www,height=\hhh]{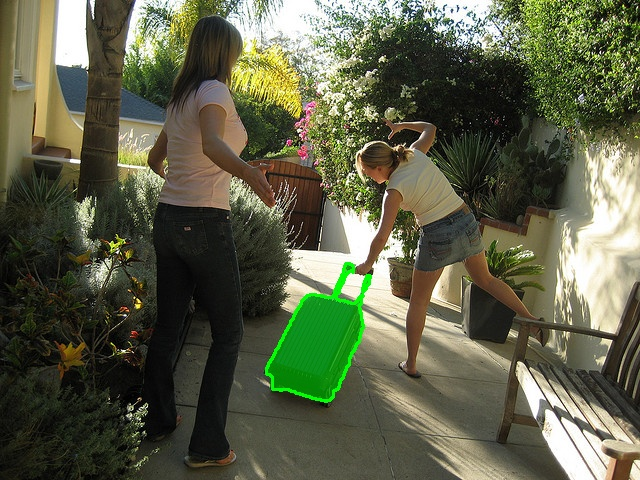}&
      \includegraphics[width=\www,height=\hhh]{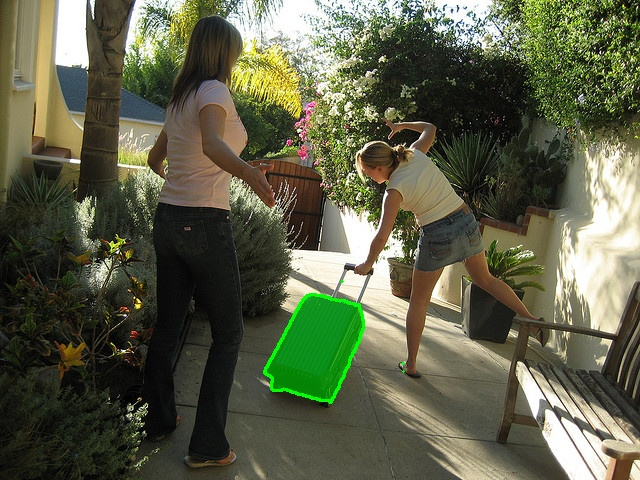}&
      \includegraphics[width=\www,height=\hhh]{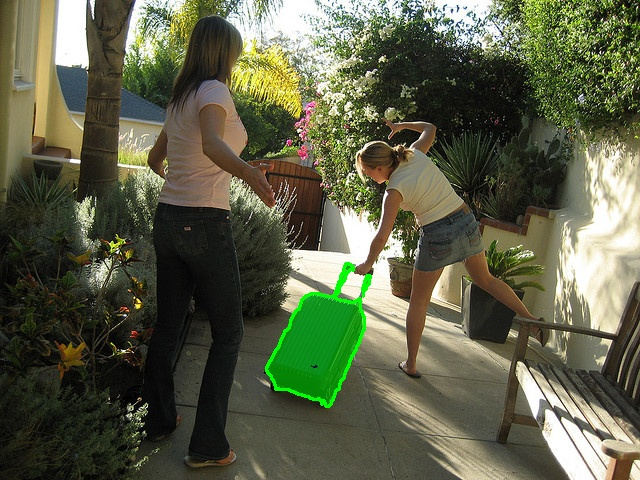}      \\[-0.25ex]
      (b) Ground truth &
      (f) \textbf{SANSA} (DISP w/o LLM) &
      (g) \textbf{SANSA} (DISP) & 
      (h) \textbf{SANSA} (EXSP) \\[1ex] 
      \multicolumn{4}{p{0.9\textwidth}}{Prompt:\texttt{\footnotesize "Segment the black regular object, with diagonal line patterns and three connected elongated gray tubular subparts on the top right".}}
      \\\\
      \includegraphics[width=\www,height=\hhh]{imgs/paper-figs/qualitative-imgs5/toilets-og.png}&
      \includegraphics[width=\www,height=\hhh]{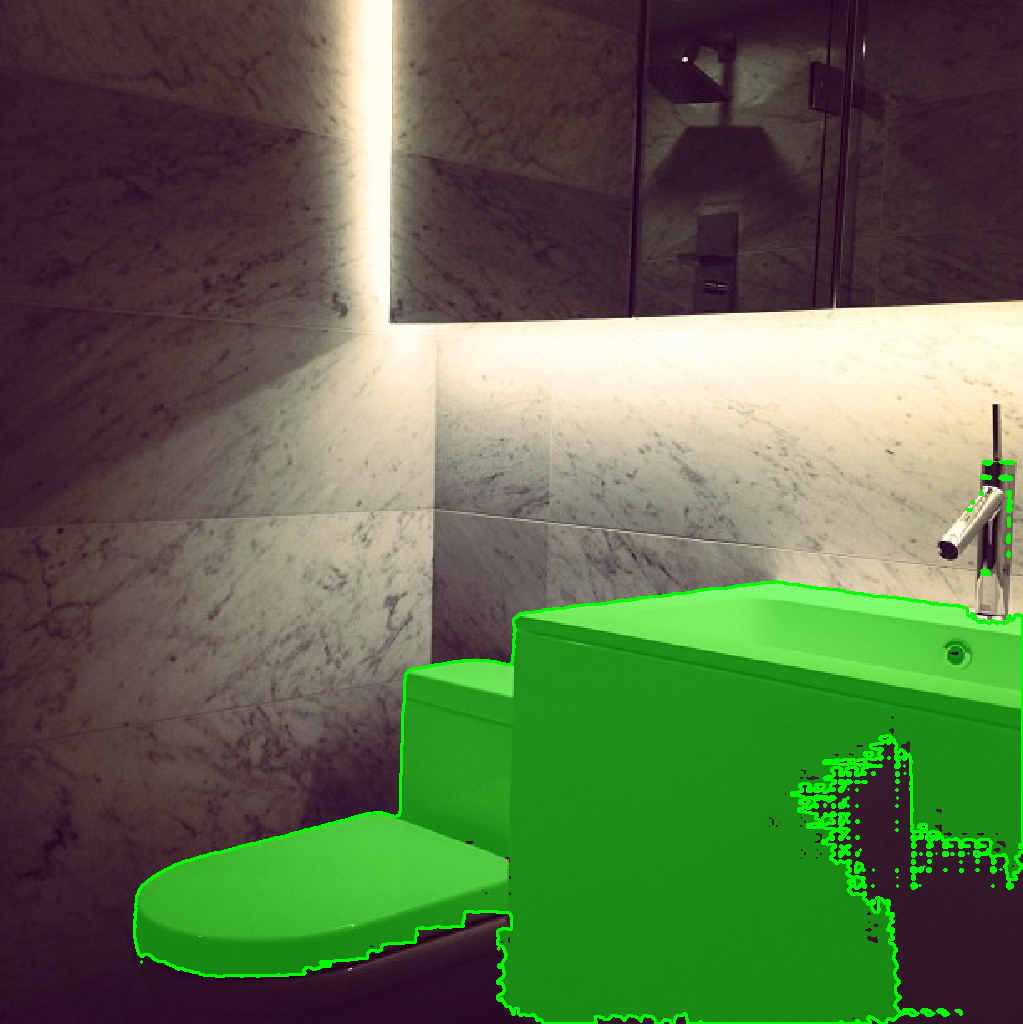}&
      \includegraphics[width=\www,height=\hhh]{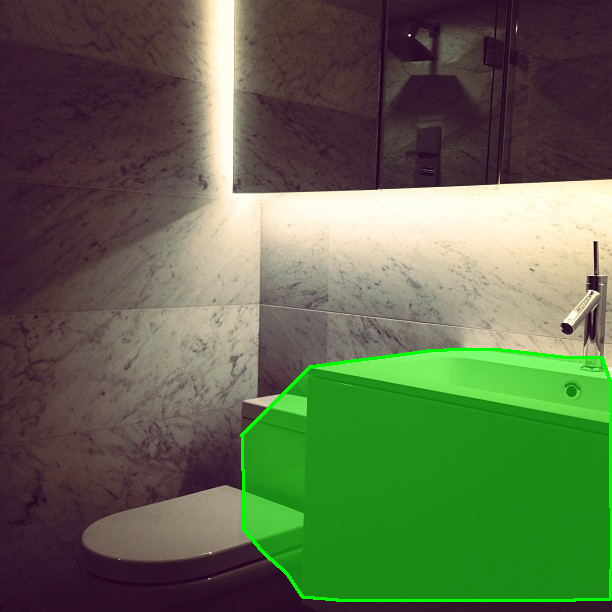}&
      \includegraphics[width=\www,height=\hhh]{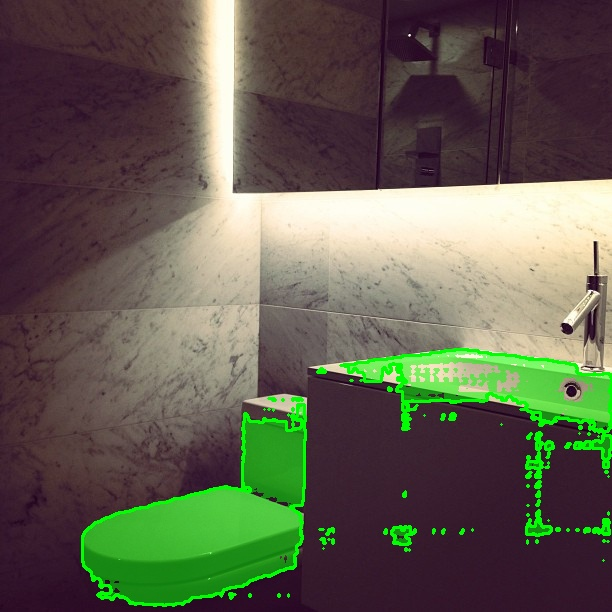}      \\[-0.25ex]
      (a) Image & 
      (c) GSVA \cite{xia2024gsva} &
      (d) PolyFormer \cite{liu2023polyformer} &
      (e) LISA \cite{lai2024lisa} \\[0.75ex]
      \includegraphics[width=\www,height=\hhh]{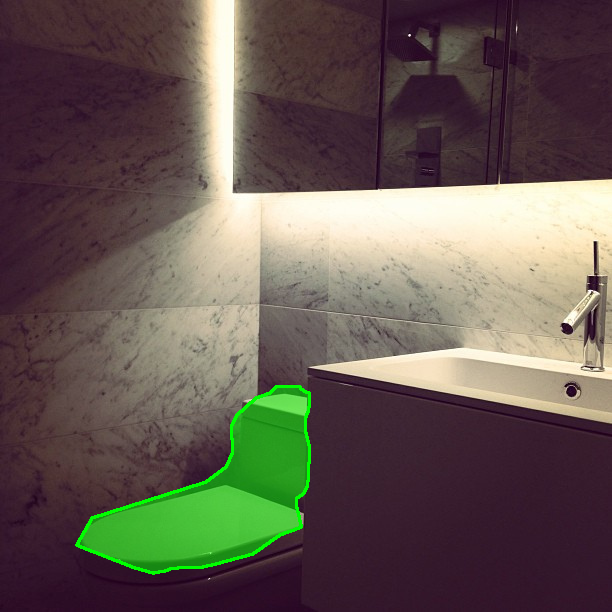}&
      \includegraphics[width=\www,height=\hhh]{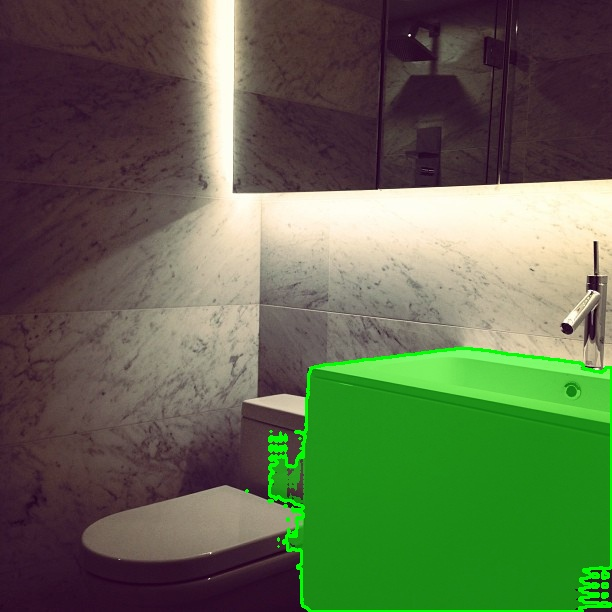}&
      \includegraphics[width=\www,height=\hhh]{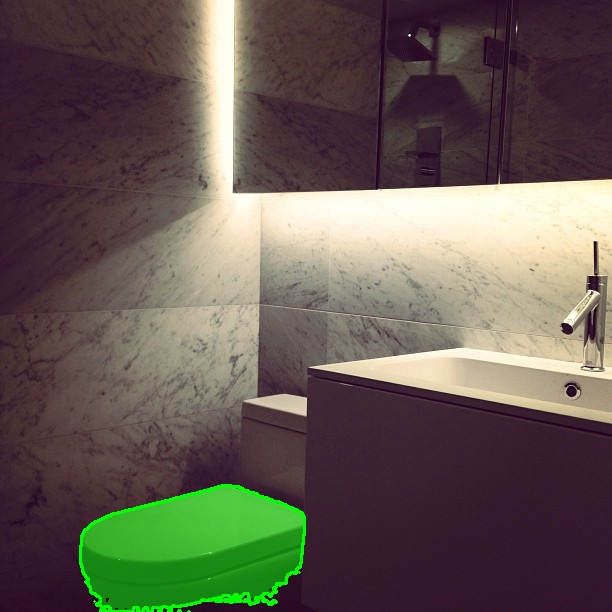}&
      \includegraphics[width=\www,height=\hhh]{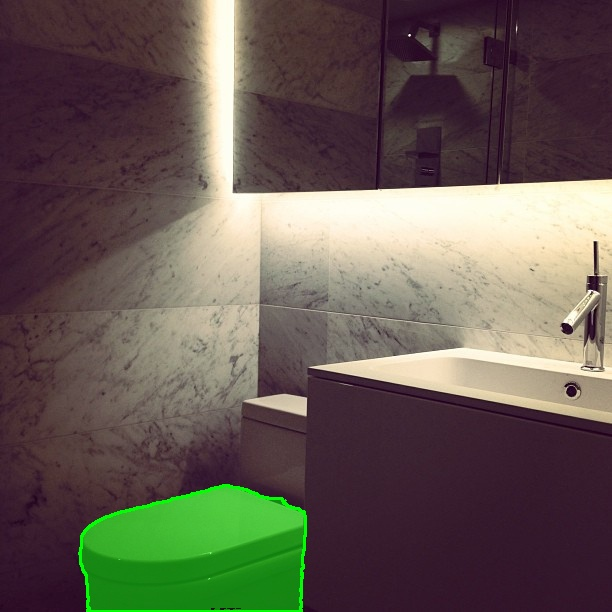}      \\[-0.25ex]
      (b) Ground truth &
      (f) \textbf{SANSA} (DISP w/o LLM) &
      (g) \textbf{SANSA} (DISP) & 
      (h) \textbf{SANSA} (EXSP) \\[1ex] 
            \multicolumn{4}{p{0.9\textwidth}}{Prompt:\texttt{\footnotesize "Segment the white object, lit by the top left angle. The surface is soft and highly reflective. The left part of the object is circular while the right part is rectangular and separated by a horizontal black line".}}
\\\\
      \includegraphics[width=\www,height=\hhh]{imgs/paper-figs/qualitative-imgs5/cat-og.png}&
      \includegraphics[width=\www,height=\hhh]{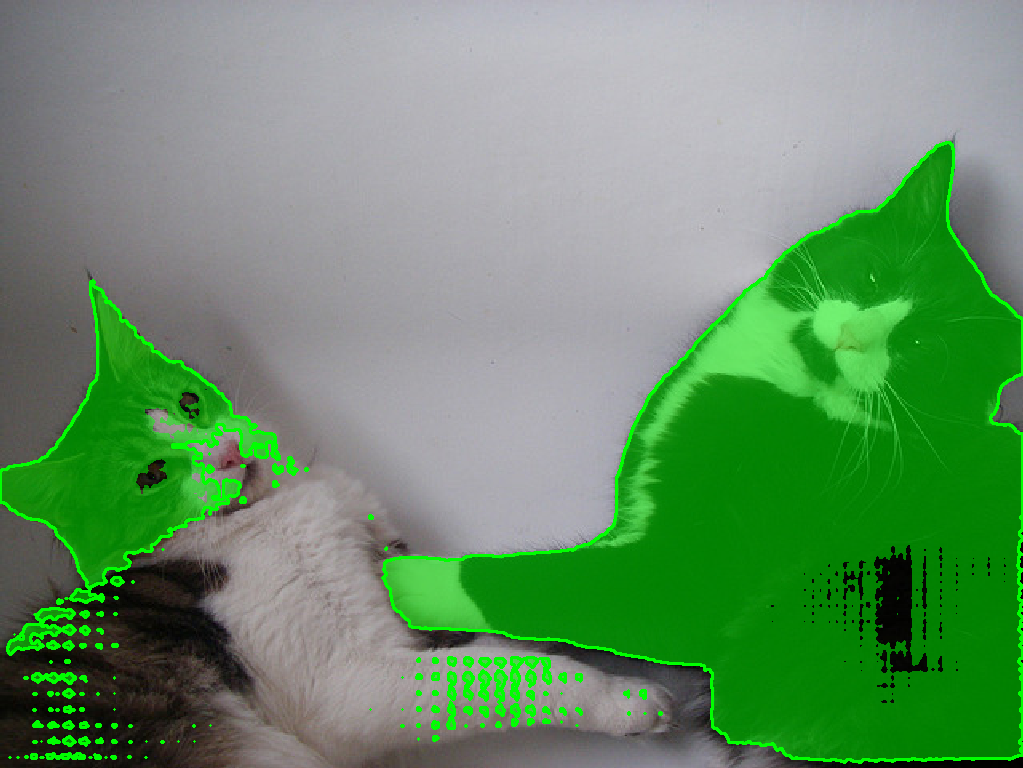}&
      \includegraphics[width=\www,height=\hhh]{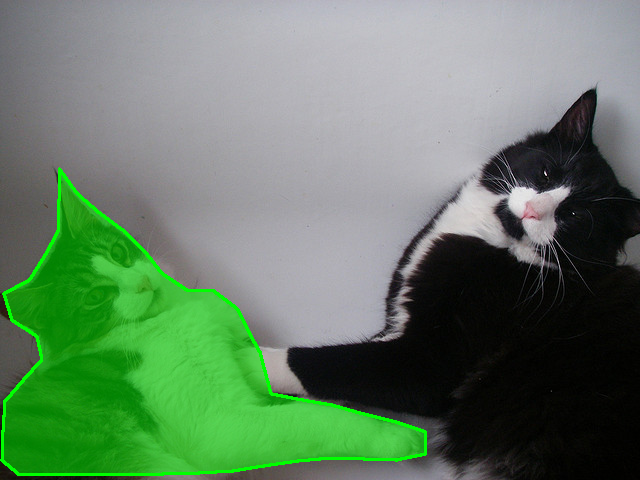}&
      \includegraphics[width=\www,height=\hhh]{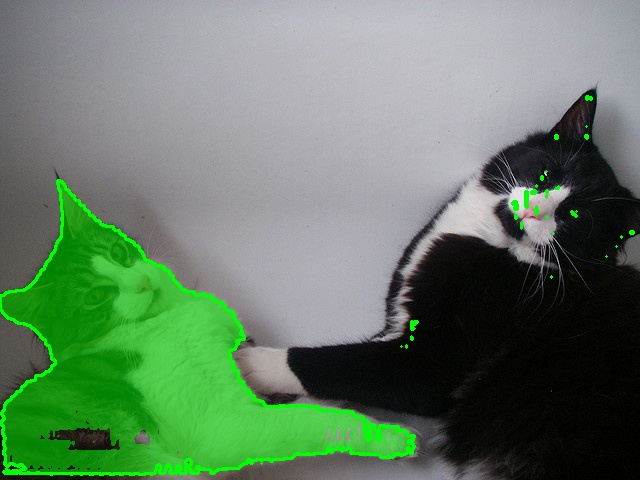}      \\[-0.25ex]
      (a) Image & 
      (c) GSVA \cite{xia2024gsva} &
      (d) PolyFormer \cite{liu2023polyformer} &
      (e) LISA \cite{lai2024lisa} \\[0.75ex]
      \includegraphics[width=\www,height=\hhh]{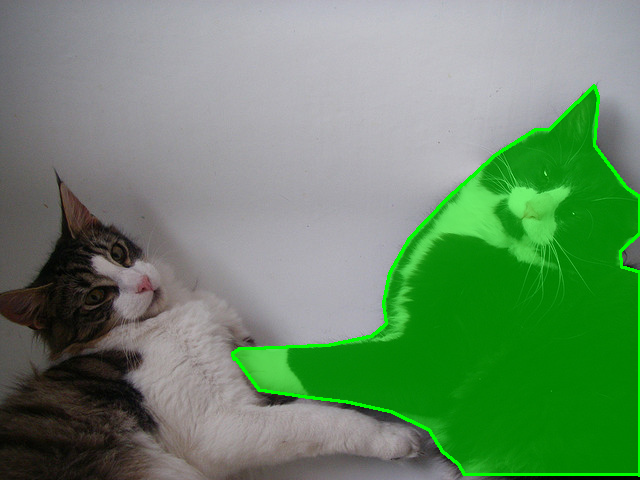}&
      \includegraphics[width=\www,height=\hhh]{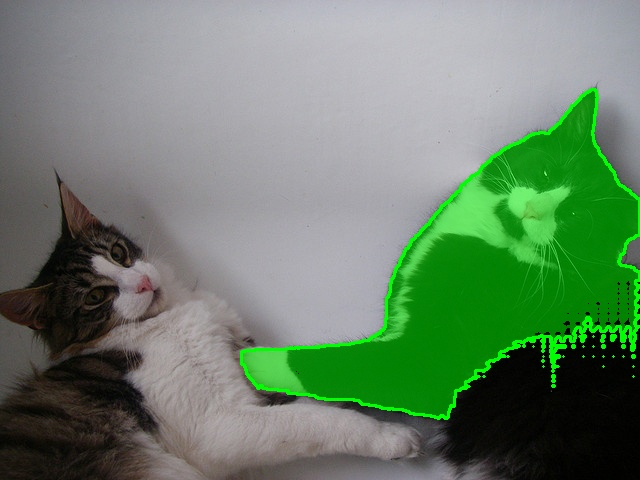}&
      \includegraphics[width=\www,height=\hhh]{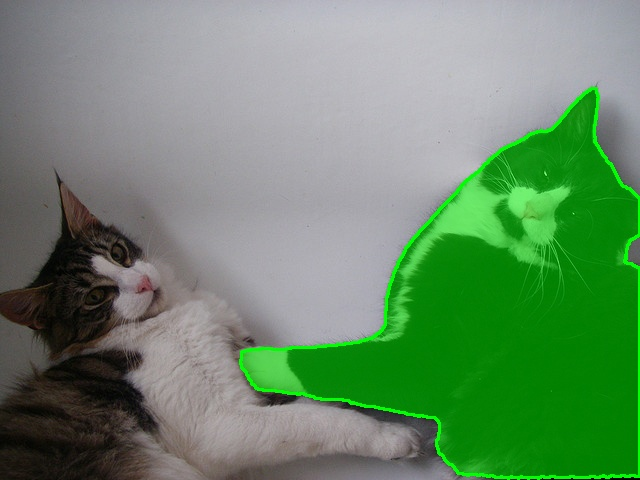}&
      \includegraphics[width=\www,height=\hhh]{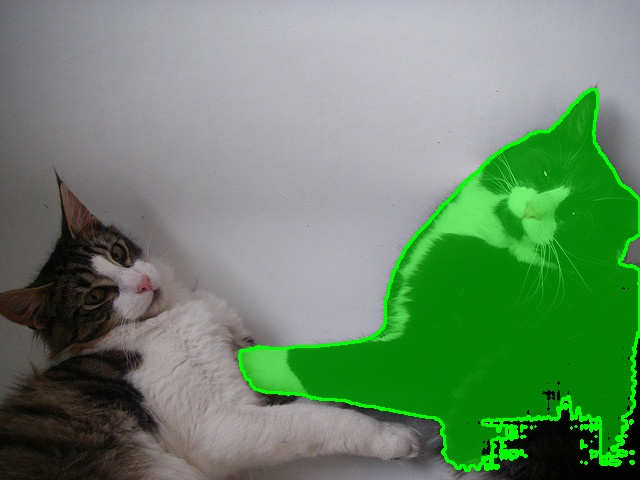}      \\[-0.25ex]
      (b) Ground truth &
      (f) \textbf{SANSA} (DISP w/o LLM) &
      (g) \textbf{SANSA} (DISP) & 
      (h) \textbf{SANSA} (EXSP) \\[1ex] 
            \multicolumn{4}{p{0.9\textwidth}}{Prompt:\texttt{\footnotesize "Segment the black and white object with smooth texture, overall with oval shape, pointy triangle on top and elongated part on the left. It also has a pink spot with thin white curved lines coming out of it".}}
      \end{tabular}
}
\caption{\textbf{{Additional qualitative segmentation results}} for (a) input image and (b) object ground truth, described by a semantic-agnostic prompt from the HP test set:
    We compare (c)-(e) pretrained segmentation VLMs and (f)-(h) our finetuned SANSA models from LISA.}
\label{fig:additional-qualitative}
\end{figure*}

\FloatBarrier
\clearpage

\bibliographystyle{IEEEtran}
\bibliography{refs}

@article{zhang2024convolutional,
  title={Convolutional neural networks rarely learn shape for semantic segmentation},
  author={Zhang, Yixin and Mazurowski, Maciej A},
  journal={Pattern Recognition},
  year={2024},
}

@inproceedings{rouge2024ccdice,
  title={{ccDice}: A topology-aware Dice score based on connected components},
  author={Roug{\'e}, Pierre and Merveille, Odyss{\'e}e and Passat, Nicolas},
  booktitle={MICCAI Workshop on Topology-and Graph-Informed Imaging Informatics},
  year={2024}
}

@article{vaswani2017attention,
  title={Attention is all you need},
  author={Vaswani, Ashish and Shazeer, Noam and Parmar, Niki and Uszkoreit, Jakob and Jones, Llion and Gomez, Aidan N and Kaiser, {\L}ukasz and Polosukhin, Illia},
  journal={NeurIPS},
  year={2017}
}

@inproceedings{dosovitskiy2021image,
 author = {Dosovitskiy, Alexey and Beyer, Lucas and Kolesnikov, Alexander and Weissenborn, Dirk and Zhai, Xiaohua and Unterthiner, Thomas and Dehghani, Mostafa and Minderer, Matthias and Heigold, Georg and Gelly, Sylvain and Uszkoreit, Jakob and Houlsby, Neil},
 title = {An Image is Worth 16x16 Words: Transformers for Image Recognition at Scale},
 year = {2021},
 booktitle = {ICLR}
}

@inproceedings{lai2024lisa,
  title={{LISA}: Reasoning segmentation via large language model},
  author={Lai, Xin and Tian, Zhuotao and Chen, Yukang and Li, Yanwei and Yuan, Yuhui and Liu, Shu and Jia, Jiaya},
  booktitle={CVPR},
  year={2024}
}

@inproceedings{
    wang2025segllm,
    title={Seg{LLM}: Multi-round Reasoning Segmentation with Large Language Models},
    author={XuDong Wang and Shaolun Zhang and Shufan Li and Kehan Li and Konstantinos Kallidromitis and Yusuke Kato and Kazuki Kozuka and Trevor Darrell},
    booktitle={ICLR},
    year={2025}
}

@inproceedings{kirillov2023segment,
 author = {Kirillov, Alexander and Mintun, Eric and Ravi, Nikhila and Mao, Hanzi and Rolland, Chloé and Gustafson, Laura and Xiao, Tete and Whitehead, Spencer and Berg, Alexander C. and Lo, Wan-Yen and Dollár, Piotr and Girshick, Ross B.},
 title = {Segment Anything},
 year = {2023},
 booktitle = {ICCV}
}

@article{carion2025sam,
  title={{SAM} 3: Segment anything with concepts},
  author={Carion, Nicolas and Gustafson, Laura and Hu, Yuan-Ting and Debnath, Shoubhik and Hu, Ronghang and Suris, Didac and Ryali, Chaitanya and Alwala, Kalyan Vasudev and Khedr, Haitham and Huang, Andrew and others},
  journal={arXiv:2511.16719},
  year={2025}
}

@inproceedings{yang2022lavt,
  title={{LAVT}: Language-aware vision transformer for referring image segmentation},
  author={Yang, Zhao and Wang, Jiaqi and Tang, Yansong and Chen, Kai and Zhao, Hengshuang and Torr, Philip HS},
  booktitle={CVPR},
  year={2022}
}

@inproceedings{xia2024gsva,
  title={{GSVA}: Generalized segmentation via multimodal large language models},
  author={Xia, Zhuofan and Han, Dongchen and Han, Yizeng and Pan, Xuran and Song, Shiji and Huang, Gao},
  booktitle={CVPR},
  year={2024}
}

@inproceedings{rasheed2024glamm,
  title={{GLaMM}: Pixel grounding large multimodal model},
  author={Rasheed, Hanoona and Maaz, Muhammad and Shaji, Sahal and Shaker, Abdelrahman and Khan, Salman and Cholakkal, Hisham and Anwer, Rao M and Xing, Eric and Yang, Ming-Hsuan and Khan, Fahad S},
  booktitle={CVPR},

  year={2024}
}

@inproceedings{chen2024sam4mllm,
  title={{SAM4MLLM}: Enhance multi-modal large language model for referring expression segmentation},
  author={Chen, Yi-Chia and Li, Wei-Hua and Sun, Cheng and Wang, Yu-Chiang Frank and Chen, Chu-Song},
  booktitle={ECCV},
  year={2024}
}

@article{chen2024expanding,
  title={Expanding performance boundaries of open-source multimodal models with model, data, and test-time scaling},
  author={Chen, Zhe and Wang, Weiyun and Cao, Yue and Liu, Yangzhou and Gao, Zhangwei and Cui, Erfei and Zhu, Jinguo and Ye, Shenglong and Tian, Hao and Liu, Zhaoyang and others},
  journal={arXiv:2412.05271},
  year={2024}
}

@inproceedings{liu2023polyformer,
  title={{Polyformer: Referring image segmentation as sequential polygon generation}},
  author={Liu, Jiang and Ding, Hui and Cai, Zhaowei and Zhang, Yuting and Satzoda, Ravi Kumar and Mahadevan, Vijay and Manmatha, R},
  booktitle={CVPR},
  year={2023}
}

@article{jiang2023mistral7b,
      title={{Mistral 7B}}, 
      author={Albert Q. Jiang and Alexandre Sablayrolles and Arthur Mensch and Chris Bamford and Devendra Singh Chaplot and Diego de las Casas and Florian Bressand and Gianna Lengyel and Guillaume Lample and Lucile Saulnier and Lélio Renard Lavaud and Marie-Anne Lachaux and Pierre Stock and Teven Le Scao and Thibaut Lavril and Thomas Wang and Timothée Lacroix and William El Sayed},
      year={2023},
    journal={arXiv:2310.06825},
}

@inproceedings{hu2022lora,
  title={{LoRA}: Low-rank adaptation of large language models},
  author={Hu, Edward J and Shen, Yelong and Wallis, Phillip and Allen-Zhu, Zeyuan and Li, Yuanzhi and Wang, Shean and Wang, Lu and Chen, Weizhu and others},
  booktitle={ICLR},
  year={2022}
}

@article{yao2024minicpmv,
  title={{MiniCPM-V: A GPT-4V level MLLM on your phone}},
  author={Yao, Yuan and Yu, Tianyu and Zhang, Ao and Wang, Chongyi and Cui, Junbo and Zhu, Hongji and Cai, Tianchi and Li, Haoyu and Zhao, Weilin and He, Zhihui and Chen, Qianyu and Zhou, Huarong and Zou, Zhensheng and Zhang, Haoye and Hu, Shengding and Zheng, Zhi and Zhou, Jie and Cai, Jie and Han, Xu and Zeng, Guoyang and Li, Dahai and Liu, Zhiyuan and Sun, Maosong},
  journal={arXiv:2408.01800},
  year={2024},
}

@inproceedings{li2023blip,
  title={{BLIP-2}: Bootstrapping language-image pre-training with frozen image encoders and large language models},
  author={Li, Junnan and Li, Dongxu and Savarese, Silvio and Hoi, Steven},
  booktitle={ICML},
  year={2023},
}

@inproceedings{liu2023improved,
  title={Improved Baselines with Visual Instruction Tuning}, 
  author={Haotian Liu and Chunyuan Li and Yuheng Li and Yong Jae Lee},
  booktitle={CVPR},
  year={2024},
}

@inproceedings{liu2023visual,
  title={Visual instruction tuning},
  author={Liu, Haotian and Li, Chunyuan and Wu, Qingyang and Lee, Yong Jae},
  booktitle={NeuRIPS},
  year={2023}
}

@inproceedings{lin2014microsoft,
  title={Microsoft {COCO}: Common objects in context},
  author={Lin, Tsung-Yi and Maire, Michael and Belongie, Serge and Hays, James and Perona, Pietro and Ramanan, Deva and Doll{\'a}r, Piotr and Zitnick, C Lawrence},
  booktitle={ECCV},
  year={2014},
}

@inproceedings{gupta2019lvis,
  title={{LVIS}: A dataset for large vocabulary instance segmentation},
  author={Gupta, Agrim and Dollar, Piotr and Girshick, Ross},
  booktitle={CVPR},
  year={2019}
}

\end{document}